\documentclass{article}

\PassOptionsToPackage{sort&compress, numbers}{natbib}
\usepackage[final]{nips}
\usepackage[utf8]{inputenc} 
\usepackage[normalem]{ulem}
\usepackage[T1]{fontenc}    
\usepackage{hyperref}       
\usepackage{url}            
\usepackage{algorithm}
\usepackage{graphicx}
\usepackage{color}
\usepackage{algorithmic}
\usepackage{eqparbox}

\usepackage{amsmath,amssymb}
\DeclareMathOperator*{\argmax}{argmax}
\usepackage{booktabs}       
\usepackage{amsfonts}       
\usepackage{nicefrac}       
\usepackage{microtype}      
\usepackage{xcolor}
\usepackage{xspace}
\usepackage{verbatim}
\usepackage{graphicx}
\usepackage{subcaption}

\newcommand{\ks}[1]{\comments{\textcolor{purple}{[Ken: #1]}}}

\definecolor{maroon}{rgb}{.5,0,0}
\definecolor{deepForestGreen}{rgb}{.2,.8,.04}
\definecolor{magenta}{rgb}{0.7,0,1}

\newcommand{\comments}[1]{#1}

\title{
Paired Open-Ended Trailblazer (POET): Endlessly Generating Increasingly Complex and Diverse Learning Environments and Their Solutions}

\author{
\textbf{Rui Wang} \hspace{5mm} \textbf{Joel Lehman} \hspace{5mm}  \textbf{Jeff Clune*}  \hspace{5mm}  \textbf{Kenneth O. Stanley*}\\
  Uber AI Labs\\
  San Francisco, CA 94103 \\
  \texttt{{ruiwang,joel.lehman,jeffclune,kstanley}@uber.com} \\
  *co-senior authors
}

\begin{document}

\maketitle
\begin{abstract}
While the history of machine learning so far largely
encompasses a series of problems posed by researchers and algorithms that learn their solutions, an important question is whether the problems themselves can be generated by the algorithm at the same time as they are being solved.  Such a process would in effect build its own diverse and expanding curricula, and the solutions to problems at various stages would become stepping stones towards solving even more challenging problems later in the process.  The \emph{Paired Open-Ended Trailblazer} (POET) algorithm introduced in this paper does just that: it pairs the generation of environmental challenges and the optimization of agents to solve those challenges.  It simultaneously explores many different paths through the space of possible problems and solutions and, critically, allows these stepping-stone solutions to transfer between problems if better, catalyzing innovation. The term \emph{open-ended} signifies the intriguing potential for algorithms like POET to continue to create novel and increasingly complex capabilities without bound.  We test POET in a 2-D bipedal-walking obstacle-course domain in which POET can modify the types of challenges and their difficulty. At the same time, a neural network controlling a biped walker is optimized for each environment.  The results show that POET produces a diverse range of sophisticated behaviors that solve a wide range of environmental challenges, many of which cannot be solved by direct optimization alone, or even through a direct-path curriculum-building control algorithm introduced to highlight the critical role of open-endedness in solving ambitious challenges.  The ability to transfer solutions from one environment to another proves essential to unlocking the full potential of the system as a whole, demonstrating the unpredictable nature of fortuitous stepping stones.  We hope that POET will inspire a new push towards open-ended discovery across many domains, where algorithms like POET can blaze a trail through their interesting possible manifestations and solutions.
\end{abstract}

\section{Introduction} \label{intro} 

Machine learning algorithms are often understood as tools for solving difficult problems.  Challenges like image classification and associated benchmarks like ImageNet \cite{Deng2009ImageNetAL} 
are chosen by people in the community as solvable and researchers focus on them to invent ever-more high-performing algorithms.  For example, ImageNet was proposed in 2009 
and modern deep neural networks such as \textit{ResNet} \cite{he:arxiv15} began to beat humans in 2015 \cite{russakovsky:arxiv14,he2:arxiv15}. 
In reinforcement learning (RL) \cite{sutton1998reinforcement}, learning to play Atari was proposed in 2013 \cite{bellemare2013arcade}
and in recent years deep reinforcement learning (RL) algorithms reached and surpassed human play across a wide variety of Atari games \cite{mnih2015,espeholt2018impala,horgan2018distributed,hessel2018multi,kapturowski2018recurrent,goexplore_blog}.
The ancient game of Go has been a challenge in AI for decades, and has been the subject of much recent research: AlphaGo and its variants can now reliably beat the world's best human professional Go players \cite{SilverHuangEtAl16nature, silver2017mastering, silver2018general}.
Each such achievement represents the pinnacle of a long march of increasing performance and sophistication aimed at solving the problem at hand.

As compelling as this narrative may be, it is not the only conceivable path forward.  While the story so far relates a series of challenges that are conceived and conquered by the community algorithm by algorithm, in a more exotic alternative 
only rarely discussed (e.g.\ \cite{forestier2017intrinsically,schmidhuber2013powerplay}), the job of the
algorithm could be to conceive \emph{both} the challenges and the solutions at the same time.  Such a process offers the novel possibility that the march of progress, guided so far by a sequence of problems
conceived by humans, could lead \emph{itself} forward, pushing the boundaries of performance autonomously and indefinitely.  In effect, such an algorithm could continually invent new environments that pose novel problems just hard enough to challenge current capabilities, but not so hard that all gradient is lost.  The environments need not arrive in a strict sequence either; they can be invented in parallel and asynchronously, in an ever-expanding tree of diverse challenges and their solutions.  

We might want such an \emph{open-ended} \cite{standish:ijcia03,langdon:metas05,bedau:alife08, ken:oriley17,taylor:alife16} 
process because the chain that leads from the capabilities of machine learning today to e.g.\ general human-level intelligence could stretch across vast and inconceivable paths of stepping stones.  There are so many directions we could go, and so many problems we could tackle, that the curriculum that leads from here to the farthest reaches of AI is beyond the scope of our present imagination.  Why then should we not explore algorithms that \emph{self-generate} their own such curricula?  If we could develop and refine such algorithms, we might find a new approach to progress that relies less upon our own intuitions about the right stepping stones and more upon the power of automation.

In fact, the only process ever actually to achieve intelligence at the human level, natural evolution, \emph{is} just such a self-contained and open-ended curriculum-generating process.  Both the problems of life, such as reaching and eating the leaves of trees for nutrition,
and the solutions, such as giraffes, are the products of the same open-ended process.  And this process unfolds not as a single linear progression,
but rather involves innumerable parallel and interacting branches radiating for more than a billion years (and is still going).  Nature is also a compelling inspiration because it has avoided convergence or stagnation, and continues to produce novel artifacts for beyond those billion years.  An intriguing question is whether it is possible to conceive an algorithm whose results would be worth waiting a billion years to see.  Open-ended algorithms in their most grandiose realization would offer this possibility.  

While a small community within the field of artificial life \cite{standish:ijcia03,langdon:metas05, lehman:alife08,graening:ppsn10,soros:alife14, soros2:alife14, brant:mcc17, ken:oriley17,ray:alife91,taylor:alife16}
has studied the prospects of open-ended computation for many years, in machine learning the prevailing assumption that algorithms should be \emph{directed} (e.g.\ towards the performance objective) has pervaded the development of algorithms and their evaluation for many years. Yet this explicitly directed paradigm has begun to soften in recent years. Researchers have begun to recognize that modern learning algorithms' hunger for data presents a long-term problem:  the data available for progress diminishes as the appropriate tasks for leveling up AI competencies increase in complexity. 
This recognition is reflected in systems based on \emph{self-play} (which is related to \emph{coevolution} \cite{ficici:alife98, wiegand:gecco01,popovici:hnc12}), such as competitive two-player RL competitions \cite{selfplay:arxiv18,silver2017mastering} wherein the task is a function of the competition, which is itself changing over time. Generative adversarial networks (GANs) \cite{goodfellow_generative_2014}, 
in which networks interact as adversaries, similarly harness a flavor of self-play and coevolution \cite{moran:arxiv18} 
to generate both challenges and solutions to those challenges.

While not explicitly motivated by open-endedness, recognition of the importance of self-generated curricula is also reflected in recent advances in automatic curriculum learning for RL, where the intermediate goals of curricula are automatically generated and selected. A number of such approaches to automatic curriculum building have been proposed.  For example, the curriculum can be produced by a generator network optimized using adversarial training to produce tasks at the appropriate level of difficulty for the agent \cite{florensa2018automatic}, by applying noise in action space to generate a set of feasible start states increasingly far from the goal (i.e.\ a reverse curriculum) \cite{pmlr-v78-florensa17a}, or through intrinsically motivated goal exploration processes (IMGEPs) \cite{forestier2017intrinsically}. The POWERPLAY algorithm offers an alternative approach to formulating continual, concurrent searches for new tasks through an increasingly general problem solver \cite{schmidhuber2013powerplay}. In Teacher-Student Curriculum Learning \cite{matiisen2017teacher},  the teacher automatically selects sub-tasks that offer the highest slope of learning curve for the student. Ideas from the field of \emph{procedural content generation} (PCG) \cite{togelius:ieeetciaig11,shaker2016procedural}, where methods are developed to generate diverse sets of levels and other contents for games, can also generate progressive curricula that help improve generality of deep RL agents \cite{justesen2017procedural}.    

The ultimate implication of this shift towards less explicitly-directed processes is to tackle the full grand challenge of open-endedness: Can we ignite a process that on its own unboundedly produces increasingly diverse and complex challenges at the same time as solving them?  And, assuming computation is sufficient, can it last (in principle) forever?  

The \emph{paired open-ended trailblazer} (POET) algorithm introduced in this paper aims to confront open-endedness directly, by evolving a set of diverse and increasingly complex environmental challenges at the same time as collectively optimizing their solutions.  A key opportunity afforded by this approach is to attempt \emph{transfers} among different environments.  That is, the solution to one environment might be a stepping stone to a new level of performance in another, which reflects our uncertainty about the stepping stones that trace the ideal curriculum to any given skill.  By radiating many 
paths of increasing challenge simultaneously, a vast array of potential stepping stones emerges from which progress in any direction might originate.  POET owes its inspiration to recent algorithms such a minimal criterion coevolution (MCC) \cite{brant:mcc17}, which shows that environments and solutions can effectively co-evolve, novelty search with local competition \cite{lehman:gecco11}, MAP-Elites \cite{mouret2015illuminating,cully:nature15}, Innovation Engines \cite{nguyen2016innovative}, which exploit the opportunity to transfer high-quality solutions from one objective among many to another, and the 
combinatorial multi-objective evolutionary algorithm (CMOEA), which extends the Innovation Engine to combinatorial tasks \cite{huizinga2016does, Joost:arxiv18}.
POET in effect combines the ideas from all of these approaches to yield a new kind of open-ended algorithm that optimizes, complexifies, and diversifies as long as the environment space and available computation will allow.

In this initial
introduction of POET, it is evaluated in a simple 2-D bipedal walking obstacle-course domain in which the form of the terrain is evolvable, from a simple flat surface to heterogeneous environments of gaps, stumps and rough terrain.  The results establish that (1) solutions found by POET for challenging environments cannot be found directly on those same environmental challenges by optimizing on them only from scratch; 
(2) neither can they be found through a curriculum-based process aimed at gradually building up to the same challenges POET invented and solved;
(3) periodic transfer attempts of solutions from some environments to others--also known as ``goal switching'' \cite{nguyen2016innovative}--is important for POET's success; (4) a diversity of challenging environments are both invented and solved in the same single run.
While the 2-D domain in this paper offers an initial hint at the potential of the approach for achieving open-ended discovery in a single run, it will be exciting to observe its application to more ambitious problem spaces in the future.  

The paper begins with background on precedent for open-endedness and POET, turning next to the main algorithmic approach.  It finally introduces the 2-D obstacle-course domain and presents both qualitative an quantitative results.

\section{Background} \label{prelim}

This section first reviews several approaches that inform the idea behind POET of tracking and storing stepping stones, which is common in evolutionary approaches to behavioral diversity. It then considers a recently proposed open-ended coevolutionary framework based on the concept of minimal criteria and, lastly, evolution strategies (ES) \cite{es}, 
which serves as the optimization engine behind POET in this paper (though in principle other reinforcement learning algorithms could be substituted in the future).  
\subsection{Behavioral Diversity and Stepping Stones}
\label{CMOEA overview}

A common problem of many stochastic optimization and search algorithms is becoming trapped on local optima, which prevents the search process from leaving sub-optimal points and reaching better ones. In the context of an open-ended search, becoming trapped is possible in more than one way: In one type of failure, the domains or problems evolving over the course of search could stop increasing their complexity and thereby stop becoming increasingly interesting.  Alternatively, the simultaneously-optimizing \emph{solutions} could be stuck at sub-optimal levels and fail to solve challenges that are solvable. Either scenario can cause the search to stagnate, undermining its open-endedness. 

Population-based algorithms going back to novelty search \cite{lehman} that encourage \emph{behavioral} (as opposed to genetic) diversity \cite{cully:nature15, mouret2015illuminating,pugh:frontiers16, huizinga2016does, nguyen2016innovative, Joost:arxiv18} have proven less susceptible to local optima, and thus naturally align more closely with the idea of open-endedness as they focus on \emph{divergence} instead of \emph{convergence}. These algorithms are based on the observation that the path to a more desirable or innovative solution often involves a series of waypoints, or \emph{stepping stones}, that may \emph{not} increasingly resemble the final solution, and are not known ahead of time. Therefore, divergent algorithms reward and preserve diverse behaviors to facilitate the preservation of potential stepping stones, which could pave the way to both a solution to a particular problem of interest and to genuinely open-ended search. 

In the canonical example of \emph{novelty search} (NS)
\cite{lehman}, individuals are selected purely based on how different their behaviors are compared to an archive of individuals from previous generations. In NS, the individuals in the archive determine the \textit{novelty} of a solution, reflecting the assumption that genuinely novel discoveries are often stepping stones to further novel discoveries.  Sometimes, a chain of novel stepping stones even reaches a solution to a problem, even though it was never an explicit objective of the search.  NS was first shown effective in learning to navigate deceptive mazes and biped locomotion problems  \cite{lehman}. 

Other algorithms are designed to generate, retain, and utilize stepping stones more explicitly. To retain as much diversity as possible, quality diversity (QD) algorithms
\cite{lehman:gecco11,cully:nature15, mouret2015illuminating,pugh:frontiers16} 
keep track of many different niches of solutions that are (unlike pure NS) being optimized 
simultaneously 
and in effect try to discover stepping stones by periodically testing the performance of offspring from one niche
in other niches, a process referred to 
as \emph{goal switching} \cite{nguyen2016innovative}.

Based on this idea, an approach called the Innovation Engine \cite{nguyen2016innovative} is able to evolve a wide range of images in a single run that are recognized with high-confidence as different image classes by a high-performing deep neural network trained on ImageNet \cite{alexnet2012}. In this domain, the Innovation Engine maintains separate niches for images of each class,  including the image that performed the best in that class (i.e.\ produced the highest activation value according to the trained deep neural network classifier). It then generates variants (perturbations/offspring) of a current niche  champion and not only checks whether such offspring are higher-performing in that niche (class), but also checks whether they are higher-performing in any other niche. Interestingly, the evolutionary path to performing well in a particular class often passes through other (oftentimes seemingly unrelated) classes that ultimately serve as stepping stones to recognizable objects. In fact, it is \emph{necessary} for the search to pass through these intermediate, seemingly unrelated stages to ultimately satisfy other classes later in the run \cite{nguyen2016innovative}.  More recently, to evolve multimodal robot behaviors, the \textit{Combinatorial Multi-Objective Evolutionary Algorithm} (CMOEA) \cite{huizinga2016does, Joost:arxiv18} builds on the Innovation Engine, but defines its niches based on different \emph{combinations of subtasks}.  This idea helps it to solve both a multimodal robotics problem with six subtasks as well as a maze navigation problem with a hundred subtasks, highlighting the advantage of maintaining a diversity of pathways and stepping stones through the search space when a metric for rewarding the most promising path is not known.  POET will similarly harness goal-switching within divergent search.

While the general ideas of promoting diversity and preserving stepping stones grew up initially within the field of evolutionary computation, they are nevertheless generic and applicable in fields such as deep reinforcement learning (RL). For example, a recent work utilizes diversity maximization to acquire complex skills with minimal supervision, which improves learning efficiency when there is no reward function. The results then become the foundation for imitation
learning and hierarchical RL \cite{eysenbach:arxiv18}. 
To address the problem of reward sparsity, \citet{savinov:arxiv18} extend the archive-based mechanism from NS that determines novelty in behaviors to determine novelty in the observation space and then to formulate the novelty bonus as an auxiliary reward. Most recently, the Go-Explore algorithm by Ecoffet et al. \cite{goexplore_blog} 
represents a new twist on QD algorithms and is designed to handle extremely sparse and/or deceptive ``hard exploration'' problems in RL. It produced dramatic improvements over the previous state of the art on the notoriously challenging hard-exploration domains of Montezuma's Revenge and Pitfall. Go-Explore maintains an archive of interestingly different stepping stones (e.g. states of a game) that have been reached so far and how to reach them. It then returns to and explores from these stepping stones to find new stepping stones, and repeats the process until a sufficiently high-quality solution is found.

\subsection{Open-Ended Search via Minimal Criterion Coevolution (MCC)}
\label{mcc overview}
Despite the recent progress of diversity-promoting algorithms such as NS and QD, they remain far from matching a genuinely open-ended search in the spirit of nature. 
One challenge in such algorithms is that although they provide pressure for ongoing divergence
in the solution space, the \emph{environment} itself remains static, limiting the scope
of what can be found in the long run.
In this spirit, one fundamental force for driving open-endedness could come from \emph{coevolution} \cite{popovici:hnc12}, which means that different individuals in a population interact with each other while they are evolving. Such multi-agent interactions have recently proven important and beneficial for systems with coevolutionary-like dynamics such as generative adversarial networks (GANs) \cite{goodfellow_generative_2014}, self-play learning in board games \cite{silver2017mastering}, and competitions between reinforcement-learning robots \cite{selfplay:arxiv18}.  Because the environment in such systems now contains other changing agents, the challenges are no longer static and evolve in entanglement with the solutions. 

However, in most, if not all, coevolutionary systems, the abiotic (non-opponent-based) aspect of the environment remains fixed. No amount of co-evolution with a fixed task (e.g. pushing the other enemy off a pedestal, or beating them at Go) will produce general artificial intelligence that can write poetry, engage in philosophy, and invent math. 
Thus an important insight is that environments themselves (including ultimately the reward function) must too change to drive truly open-ended dynamics. 
Following this principle, an important predecessor to the POET algorithm in this paper is the recent \emph{minimal criterion coevolution} (MCC) 
algorithm \cite{brant:mcc17}, which explores an alternative paradigm for open-endedness through coevolution. In particular, it implements a novel coevolutionary framework that pairs a population of evolving \emph{problems} (i.e.\ environmental challenges) with a co-evolving population of \emph {solutions}.  
That enables new problems to evolve from existing ones and for new kinds of solutions to branch from existing solutions.  As problems and solutions proliferate and diverge in this ongoing coupled interaction, the potential for open-endedness arises.

Unlike conventional coevolutionary algorithms that are usually divided between competitive coevolution, where individuals are rewarded according to their performance against other individuals in a competition \cite{ficici:alife98}, and cooperative coevolution, where instead fitness is a measure of how well an individual performs in a cooperative team with other evolving individuals \cite{wiegand:gecco01}, MCC introduce a new kind of coevolution that evolves two interlocking populations whose members earn the right to reproduce by satisfying a \emph{minimal criterion} \cite{soros:alife14, soros2:alife14, lehman:gecco10a} with respect to the other population, as both populations are gradually shifting simultaneously.  For example, in an example in \citet{ brant:mcc17} with mazes (the problems) and maze solvers (the solutions, represented by neural networks), the minimal criterion for the solvers is that they must solve at least one of the mazes in the maze population, and the minimal criterion for the mazes is to be solved by at least one solver.

The result is that the mazes increase in complexity and the neural networks continually evolve to solve them. In a single run without any behavior characterization or archive (which are usually needed in QD algorithms), MCC can continually generate novel, diverse and increasingly complex mazes and maze solvers as new mazes provide new opportunities for innovation to the maze solvers (and vice versa), hinting at open-endedness.  However, in MCC there is no force for \emph{optimization} within each environment (or maze in the example experiment).  That is, once a maze is solved there is no pressure to \emph{improve} its solution; instead, it simply becomes a potential stepping stone to a solution to another maze.  As a result, while niches and solutions proliferate, there is no pressure for \emph{mastery} within each niche, so we do not see the best that is possible. Additionally, in MCC problems have to be solvable by the \emph{current} population, so complexity can only arise through drift. As an example, imagine an environmental challenge $E'$ that is similar to a currently solved problem $E$, but a bit harder such that none of the current agents in the population can solve it. Further imagine that if we took some of the current agents (e.g. the agent that solves $E$) and optimized them to solve $E'$, at least one agent could learn to solve $E'$ in a reasonable amount of time. MCC would still reject and delete $E'$ if it was generated because none of the current population solves it right away. In contrast, in POET, we accept problems if (according to some heuristics) they seem likely to be improved upon and/or solved after some amount of dedicated optimization effort (so POET will likely keep $E'$), enabling a more direct and likely swifter path to increasingly complex, yet solvable challenges.

\subsection{Evolution Strategies (ES)} \label{evolution strategies details}
In the POET implementation in this paper, ES plays the role of the optimizer (although other optimization algorithms should also work).  Inspired by natural evolution, ES \cite{rechenberg1978evolutionsstrategien} represents a broad class of population-based optimization algorithms. The method referred to here and subsequently as ``ES'' is a version of ES popularized by \citet{es} that was recently applied with large-scale deep learning architectures to modern RL benchmark problems. This version of ES draws inspiration from Section 6 of \citet{williams1992simple} (i.e.\ REINFORCE with multiparameter distributions), as well as from subsequent population-based optimization methods including Natural Evolution Strategies (NES) \cite{wierstra2008natural} and Parameter-Exploring Policy Gradients (PEPG) \cite{sehnke2010parameter}. More recent investigations have revealed the relationship of ES to finite difference gradient approximation \cite{lehman:gecco18esfd} and stochastic gradient descent \cite{zhang:arxiv17}. 

In the typical context of RL, we have an environment, denoted as $E(\cdot)$, and an agent under a parameterized policy whose parameter vector is denoted as $w$. The agent tries to maximize its reward, denoted as $E(w)$, as it interacts with the environment. In ES, $E(w)$ represents the stochastic reward experienced over a full episode of an agent interacting with the environment. Instead of directly optimizing $w$ to maximize $E(w)$, ES seeks to maximize the expected fitness over a population of $w$, $J(\theta) = \mathbb{E}_{w \sim p_{\theta}(w)}[E(w)]$, where $w$ is sampled from a probability distribution $p_{\theta}(w)$ parameterized by $\theta$. 
Using  the log-likelihood trick \cite{williams1992simple} and estimating the expectation above over $n$ samples allows us to write the gradient of $J(\theta)$ with respect to $\theta$ as:
\begin{gather*}\label{eq:1}
\scalebox{1.}{$
\nabla_{\theta}J(\theta) \approx \frac{1}{n}\sum_{i=1}^{n}E(\theta_{i})\nabla_{\theta} \log p_{\theta}(\theta_{i}).
$}
\end{gather*}

Intuitively, this equation says that for every sample agent $\theta_i$, the higher the performance of that agent $E(\theta_i)$, the more we should move $\theta$ in the direction of the gradient that increases the likelihood of sampling that agent, which is $\nabla_{\theta} \log p_{\theta}(\theta_{i})$. 
Following the convention in \citet{es}, $\theta_{i}, i = 1, \ldots, n,$ are $n$ samples of parameter vectors drawn from an isotropic multivariate Gaussian with mean $\theta$ and a covariance $\sigma^{2},$ namely, $\mathcal{N}(\theta,\,\sigma^{2}I)$. (The covariance can be fixed or adjusted over time depending on the implementation.) Note that $\theta_{i}$ can be equivalently obtained by applying additive Gaussian noise $\epsilon_{i} \sim \mathcal{N}(0,\,I)$ to a given parameter vector $\theta$ as $\theta_{i} = \theta + \sigma\epsilon_{i}$. With that notation, the above estimation of the gradient of $J(\theta)$ with respect to $\theta$ can be written as:
\begin{gather*}\label{eq:2}
\scalebox{1.}{$
\nabla_{\theta}J(\theta) \approx \frac{1}{n\sigma}\sum_{i=1}^{n}E(\theta + \sigma\epsilon_{i})\epsilon_{i}.
$}
\end{gather*}

Algorithm \ref{alg:es_step} illustrates the calculation of one optimization step of ES, which can efficiently be parallelized over distributed workers \cite{es}.
Once calculated, the returned optimization step is added to the current parameter vector to obtain the next parameter vector. Note that implementations of ES including our own usually follow the approach of \citet{es} and rank-normalize $E(\theta + \sigma\epsilon_{i})$ before taking the weighted sum, which is a variance reduction technique.
Overall, ES has exhibited performance on par with some of the traditional, simple gradient-based RL algorithms on difficult RL domains (e.g. DQN \cite{mnih2015} and A3C \cite{a3c}), 
including Atari environments and simulated robot locomotion \cite{es, conti:nses_nips2018}.
More recently, NS and QD algorithms have been shown possible to hybridize with ES to further improve its performance on sparse or deceptive deep RL tasks, while retaining scalability \cite{conti:nses_nips2018}, 
providing inspiration for its novel hybridization within an CMOEA- and MCC-like algorithm in this paper.

\begin{algorithm}[htp]
\caption{ES\_STEP}
\label{alg:es_step}
\begin{algorithmic}[1]
   \STATE {\bfseries Input:} an agent denoted by its policy parameter vector $\theta$, an environment $E(\cdot)$,  learning rate $\alpha$, noise standard deviation $\sigma$
  
   \STATE Sample $\epsilon_{1}, \epsilon_{2}, \ldots, \epsilon_{n} \sim \mathcal{N}(0, I)$
   \STATE Compute $E_{i} = E(\theta + \sigma\epsilon_{i})$ for $i = 1, \ldots, n$
   \STATE {\bfseries Return:} $\alpha \frac{1}{n\sigma} \sum_{i=1}^{n} E_{i} \epsilon_{i}$

\end{algorithmic}
\end{algorithm}



\section{The Paired Open-Ended Trailblazer (POET) Algorithm} \label{algos} 
POET is designed to facilitate an open-ended process of discovery within a single run. It maintains a population of environments (for example, various obstacle courses) and a population 
of agents (for example, neural networks that control a robot to solve those courses), and each environment is paired with an agent to form an \textit{environment-agent pair}. POET in effect implements an ongoing divergent coevolutionary interaction 
among all its agents and environments in the spirit of MCC \cite{brant:mcc17}, but with the added goal of explicitly optimizing the behavior of each agent within its paired environment in the spirit of CMOEA \cite{huizinga2016does, Joost:arxiv18}. 
It also elaborates on the minimal criterion in MCC by aiming to maintain only those newly-generated environments that are not too hard \emph{and} not too easy for the current population of agents.
The result is a \emph{trailblazer} algorithm, one that continually forges new paths to both increasing challenges and skills
within a single run.  The new challenges are embodied by the new environments that are continually created, and the increasing skills are embodied by the neural network controllers attempting to solve each environment. Existing skills are harnessed both by optimizing agents paired with environments and by attempting to transfer current agent behaviors to new environments to identify promising stepping stones. 

The fundamental algorithm of POET is simple:  The idea is to maintain a list of active environment-agent pairs \texttt{EA\_List} that begins with a single starting pair  $(E^{\textrm{init}}(\cdot),\theta^{\textrm{init}})$, where $E^{\textrm{init}}$ is a simple environment (e.g.\ an obstacle course of entirely flat ground) and $\theta^{\textrm{init}}$ is a randomly initialized weight vector (e.g.\ for a neural network).  POET then has three main tasks that it performs at each iteration of its main loop: 

\begin{enumerate}
    \item generating new environments $E(\cdot)$ from those currently active, 
    \item optimizing paired agents within their respective environments, and 
    \item attempting to transfer current agents $\theta$ from one environment to another.  
\end{enumerate}

Generating new environments is how POET continues to produce new challenges.  To generate a new environment, POET simply mutates (i.e.\ randomly perturbs) the encoding (i.e.\ the parameter vector) of an active environment.  However, while it is easy to generate perturbations of existing environments, the delicate part is to ensure both that (1) the paired agents in the originating (parent) environments have exhibited sufficient progress to suggest that reproducing their respective environments would not be a waste of effort, and (2) when new environments are generated, they are not added to the current population of environments unless they are neither too hard nor too easy for the current population. 
Furthermore, priority is given to those candidate environments that are most novel, which produces a force for diversification that encourages many different kinds of problems to be solved in a single run.

These checks together ensure that the curriculum that emerges from adding new environments is smooth and calibrated to the learning agents.  In this way, when new environments do make it into the active population, they are genuinely stepping stones for continued progress and divergence.  The population of active environments is capped at a maximum size, and when the size of the population exceeds that threshold, the oldest environments are removed to make room (as in a queue).  That way, environments do not disappear until absolutely necessary, giving their paired agents time to optimize and allowing skills learned in them to transfer to other environments.

POET optimizes its paired agents at each iteration of the main loop.   The idea is that every agent in POET should be continually improving within its paired environment.  In the experiments in this paper, each such iteration is a step of ES, but any reinforcement learning algorithm could conceivably apply.  The objective in the optimization step is simply to maximize whatever performance measure applies to the environment (e.g.\ to walk as far as possible through an obstacle course).  The fact that each agent-environment pair is being optimized independently affords easy parallelization, wherein all the optimization steps can in principle be executed at the same time.

Finally, attempting \emph{transfer} is the ingredient that facilitates serendipitous cross-pollination: it is always possible that progress in one environment could end up helping in another.  For example, if the paired agent $\theta^A$ in environment $E^A(\cdot)$ is stuck in a local optimum, one remedy could be a transfer from the paired agent $\theta^B$ in environment $E^B(\cdot)$.  If the skills learned in the latter environment apply, it could revolutionize the behavior in the former, reflecting the fact that the most promising stepping stone to the best possible outcome may not be the current top performer in that environment \cite{stanley:book15, nguyen2016innovative,mouret2015illuminating}.  Therefore, POET continually attempts transfers among the active environments.  These transfer attempts are also easily parallelized because they too can be attempted independently.

More formally, Algorithm \ref{alg:poet} describes this complete main loop of POET. Each component is shown in pseudocode: creating new environments, optimizing each agent, and attempting transfers.  The specific implementation details for the subroutines MUTATE\_ENVS (where the new environments are created) and EVALUATE\_CANDIDATES (where transfers are attempted) are given in Supplemental Information. 


\begin{algorithm}[htp]
\caption{POET Main Loop}
\label{alg:poet}
\begin{algorithmic}[1]
   \STATE {\bfseries Input:} initial environment $E^{\textrm{init}}(\cdot)$, its paired agent denoted by policy parameter vector $\theta^{\textrm{init}}$, learning rate $\alpha$, noise standard deviation $\sigma$, iterations $T$, mutation interval $N^{\textrm{mutate}}$, transfer interval $N^{\textrm{transfer}}$
   \STATE {\bfseries Initialize:} Set \texttt{EA\_list} empty 
   \STATE Add $(E^{\textrm{init}}(\cdot)$, $\theta^{\textrm{init}})$ to \texttt{EA\_list}
   \FOR{$t=0$ {\bfseries to} $T-1$}
   \IF{$t > 0$ {\bfseries and} $t \mod N^{\textrm{mutate}} = 0$}
   \STATE $\texttt{EA\_list} =$ MUTATE\_ENVS$(\texttt{EA\_list})$ \COMMENT{\bfseries new environments created by mutation}
   \ENDIF
   \STATE $M = \textrm{len}(\texttt{EA\_list})$
   \FOR{$m=1$ {\bfseries to} $M$} 
   \STATE $E^{m}(\cdot)$, $\theta^{m}_{t}$ = \texttt{EA\_list}[m]
   \STATE $\theta^{m}_{t+1} = \theta^{m}_{t} +$ ES\_STEP$(\theta^{m}_{t}, E^{m}(\cdot), \alpha, \sigma)$ \COMMENT{\bfseries each agent independently optimized} 
   \ENDFOR
   \FOR{$m=1$ {\bfseries to} $M$} 
   \IF{$M > 1$ {\bfseries and} $t \mod N^{\textrm{transfer}} = 0$} 
   \STATE $\theta^{\textrm{top}} = $ EVALUATE\_CANDIDATES$(\theta^{1}_{t+1}$, \ldots, $\theta^{m-1}_{t+1}$, $\theta^{m+1}_{t+1}$, \ldots, $\theta^{M}_{t+1}$, $E^{m}(\cdot)$, $\alpha$, $\sigma)$
   \IF{$E^{m}(\theta^{\textrm{top}}) > E^{m}(\theta^{m}_{t+1})$}
   \STATE $\theta^{m}_{t+1} = \theta^{\textrm{top}}$ \COMMENT{\bfseries transfer attempts}
   \ENDIF
   \ENDIF
   \STATE $\texttt{EA\_list}[m] = (E^{m}(\cdot), \theta^{m}_{t+1})$  
   
   \ENDFOR
   \ENDFOR

\end{algorithmic}
\end{algorithm}

As noted above, the independence of many of the operations in POET, such as optimizing individual agents within their paired environments and attempting transfers, makes it feasible to harness the power of many processors in parallel.  In the implementation of the experiment reported here, each run harnessed 256 parallel CPU Cores.
Our software implementation of POET, which will be released as open source code shortly,
allows seamless parallelization over any number of cores: workers are managed through Ipyparallel to automatically distribute all current tasks so that requests for specific operations and learning steps can be made without concern for how they will be distributed.

Provided that a space of possible environmental challenges can be encoded, the hope is that the POET algorithm can then start simply and push outward in parallel along an increasingly challenging frontier of challenges, some of which benefit from the solutions to others.  The next section describes a platform designed to test this capability.

\section{Experiment Setup And Results} \label{experiments} 


An effective test of POET should address the hypothesis that it can yield an increasingly challenging set of environments, many with a satisfying solution, all in a single run.  Furthermore, we hope to see evidence for the benefit of cross-environment transfers.  The experimental domain in this work is a modified version of the ``Bipedal Walker Hardcore'' environment of the OpenAI Gym \cite{brockman}.  Its simplicity as a 2-D walking domain with various kinds of possible terrain makes it easy to observe and understand qualitatively different ambulation strategies  simply by viewing them.  Furthermore, the environments are easily modified, enabling numerous diverse obstacle courses to emerge to showcase the possibilities for adaptive specialization and generalization. Finally, it is relatively fast to simulate, allowing many long experiments compared to the more complex environments where we expect this algorithm to shine in the future.

\subsection{Environment and Experiment Setup}
\begin{figure}
  \centering
  \includegraphics[width=.99\linewidth]{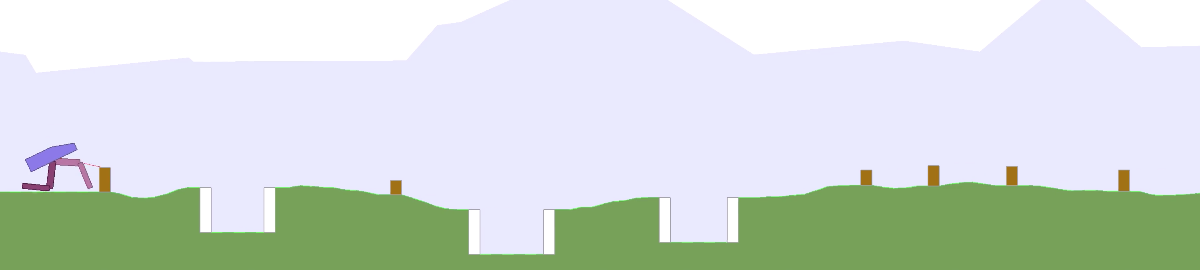}
\caption{\textbf{A landscape from the Bipedal Walker environment}. Possible obstacles include stumps, gaps, stairs, and surfaces with different amounts of roughness.}
  \label{fig:typical_landscape}
\end{figure}

The agent’s hull is supported by two legs (the agent appears on the left edge of figure \ref{fig:typical_landscape}).
The hips and knees of each leg are controlled by two motor joints, creating an action space of four dimensions. The agent has ten LIDAR rangefinders for perceiving obstacles and terrain, whose measurements are included in the state space. Another 14 state variables include hull angle, hull angular velocity, horizontal and vertical speeds, positions of joints and joint angular velocities, and whether legs touch the ground \cite{brockman}. 

Guided by its sense of the outside world through LIDAR and its internal sensors, the agent is required to navigate, within a time limit and without falling over, across an environment of a generated terrain that consists of one or more types of obstacles. These can include stumps, gaps,
and stairs on a surface with a variable degree of roughness, as illustrated in Figure \ref{fig:typical_landscape}. 
As described in the following equation, reward is given for moving forward, with almost no constraints on agents' behaviors other than that they are encouraged to keep their hulls straight and minimize motor torque. If the agent falls, the reward is $-100$. 
\[
    \text{Reward per step}= 
\begin{cases}
    -100,& \text{if robot falls }\\
    130 \times \Delta x-5\times\Delta \text{hull\_angle} - 0.00035\times \text{applied\_torque},& \text{otherwise.}
\end{cases}
\]
The episode immediately terminates when the time limit (2,000 time steps) is reached, when the agent falls, or when it completes the course. In this work, we define an environment as \emph{solved} when it both reaches the far end of the environment and obtains a score of 230 or above.
Based on our observations, the score of 230 ensures that the walker is reasonably efficient.

Following the architecture of controllers for the same domain by \citet{davidha:arxiv18},
all controllers in the experiments are implemented as neural networks with 3 fully-connected layers with $tanh$ activation  functions. The controller has 24 inputs and 4 outputs, all bounded between \text{-}1 and 1, with 2 hidden layers of 40 units each. Each ES step (Algorithm \ref{alg:es_step}) has a population size of 512 (i.e.\ the number of parameters sampled to create the weighted-average for one  gradient step is 512) and and updates weights through the Adam optimizer \cite{kingma2014adam}.
The learning rate is initially set to $0.01$, and decays to $0.001$ with a decay factor of $0.9999$ per step. The noise standard deviation for ES
is initially set to $0.1$, and decays to $0.01$ with a decay factor of $0.999$ per step. When any environment-agent pair accepts a transfer or when a child environment-agent pair is first created, we reset the state of its Adam optimizer, and set the learning rate and noise standard deviation to their initial values, respectively. 

\subsection{Environment Encoding and Mutation}

Intuitively, the system should begin with a single flat and featureless environment 
whose paired policy can be optimized easily (to walk on flat ground).  
From there, new environments will continue to be generated from their predecessors, while their paired policies are simultaneously optimized.
The hope is that a wide variety of control strategies and skill sets will allow the completion of an ever-expanding set of increasingly complex obstacle courses, all in a single run.  

To enable such a progression, we adopt a simple encoding 
to represent the search space of possible environments. 
As enumerated in Table \ref{env-mutation-table}, there are five types of obstacles that can be distributed throughout the environment: (1) stump height, (2) gap width, (3) step height, (4) step number (i.e.\ number of stairs), and (5) surface roughness. 
Three of these obstacles, e.g.\ stump height, are encoded as a pair of parameters (or \emph{genes}) that form an interval from which the actual value for each instance of that type of obstacle in a given environment is uniformly sampled.  As will be described below, in some experiments, some obstacle types are intentionally omitted, allowing us to restrict the experiment to certain types of obstacles. 
When selected to mutate for the first time, obstacle parameters are initialized to the corresponding initial values shown in Table \ref{env-mutation-table}. 
For subsequent mutations, an obstacle parameter takes a \emph{mutation step} (whose magnitude is given in Table \ref{env-mutation-table}), and either adds or subtracts the step value from its current value.
Note that for surface roughness, both the initial value and every subsequent mutation step are sampled uniformly from $(0, 0.6)$ because the impact of roughness on the agent was found to be significant sometimes even from very small differences. 
The value of any given parameter cannot exceed its \emph{maximum value}. 

Because the parameters of an environment define a distribution, the actual environment sampled from that distribution is the result of a random \emph{seed}.  This seed value is stored with each environment so that environments can be be reproduced precisely, ensuring repeatability.  (The population of many environments and the mutation of environments over time still means that training overall does not occur on only one deterministic environment.)
With this encoding, all possible environments can be uniquely defined by the values for each obstacle type in addition to the seed that is kept with the environment. 

Any environments that satisfy the reproduction eligibility condition in Line 7 of Algorithm \ref{alg:mutate_niches} (which is given in Supplemental Information) are allowed to mutate to generate a child environment. In our experiments, this condition is that the paired agent of the environment achieves a reward of 200 or above,  which generally indicates that the agent is capable of reaching the end of the terrain (though slightly below the full success criterion of 230). 
To construct \texttt{child\_list} in Line 15 in Algorithm \ref{alg:mutate_niches}, the set of eligible parents is sampled uniformly to choose a parent, which is then mutated to form a child that is added to the list.  This process is repeated until \texttt{child\_list} reaches \texttt{max\_children}, which is 512 in our experiments.  Each child is generated from its parent by independently picking and mutating some or all the available parameters of the parent environment and then choosing a new random seed.
The \emph{minimal criterion} (MC) $50 \leq E^{\textrm{child}}(\theta^{\textrm{child}}) \leq 300$ 
then filters out child environments that appear too challenging or too trivial for the current level of capability of agents. In case the number of child environments that satisfy the MC is more than the maximum number of children admitted per reproduction, those with lower $E^{\textrm{child}}(\theta^{\textrm{child}})$ are admitted until the cap is reached. 

\begin{table}[H]
\begin{center}
\begin{small}
\begin{sc}
\begin{tabular}{lccccc}
\hline
\begin{tabular}{@{}c@{}}Obstacle \\ Type\end{tabular} & \begin{tabular}{@{}c@{}}Stump \\ Height\end{tabular} & \begin{tabular}{@{}c@{}}Gap \\ Width\end{tabular} & \begin{tabular}{@{}c@{}}Step \\ Height\end{tabular} & \begin{tabular}{@{}c@{}}Step \\ Number\end{tabular} & Roughness\\

\hline
Initial Value & $(0.0, 0.4)$ & $(0.0, 0.8)$ &  $(0.0, 0.4)$ & $1$ & uniform$(0, 0.6)$\\
Mutation Step & $0.2$ & $0.4$ & $0.2$ & $1$ & uniform$(0, 0.6)$\\
Max Value & $(5.0, 5.0)$ & $(10.0, 10.0)$ & $(5.0, 5.0)$ & $9$ & $10.0$\\
\hline
\end{tabular}
\end{sc}
\end{small}
\end{center}
\caption{\textbf{Environmental parameters (genes) for the Bipedal Walker problem.} 
}\label{env-mutation-table}
\vskip -0.1in
\end{table}


\subsection{Results}

An important motivating hypothesis for POET is that the stepping stones that lead to solutions to very challenging environments are more likely to be found through a divergent, open-ended process than through a direct attempt to
optimize in the challenging environment.  
Indeed, if we take a particular environment evolved in POET and attempt to run ES (the same optimization algorithm used in POET) on that specific environment from scratch, the result is often premature convergence to degenerate behavior.
The first set of experiments focus on this phenomenon by running POET with only \emph{one} obstacle type enabled (e.g.\ gaps in one case, roughness in another, and stumps in another). 
That way, we can see that even with a single obstacle type, the challenges generated by POET (which POET solves) are too much for ES on its own. 

\begin{figure}
    \centering
    \begin{subfigure}[b]{1.0\textwidth}
        \includegraphics[width=\textwidth]{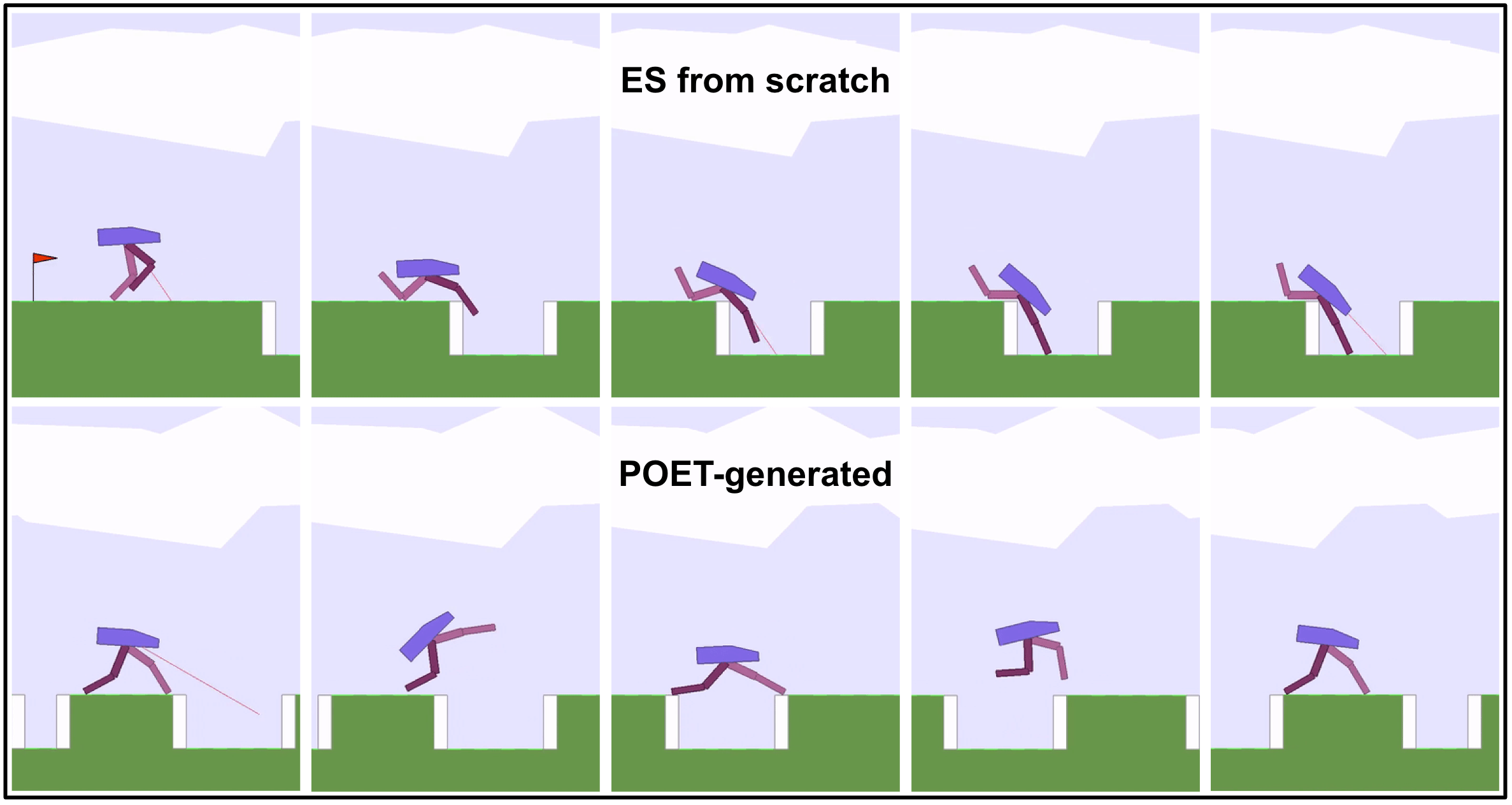}
        \caption{ES-/POET-generated agents attempting gaps}
        \label{fig:large_gap}
    \end{subfigure}
    \begin{subfigure}[b]{0.49\textwidth}
        \includegraphics[width=\textwidth]{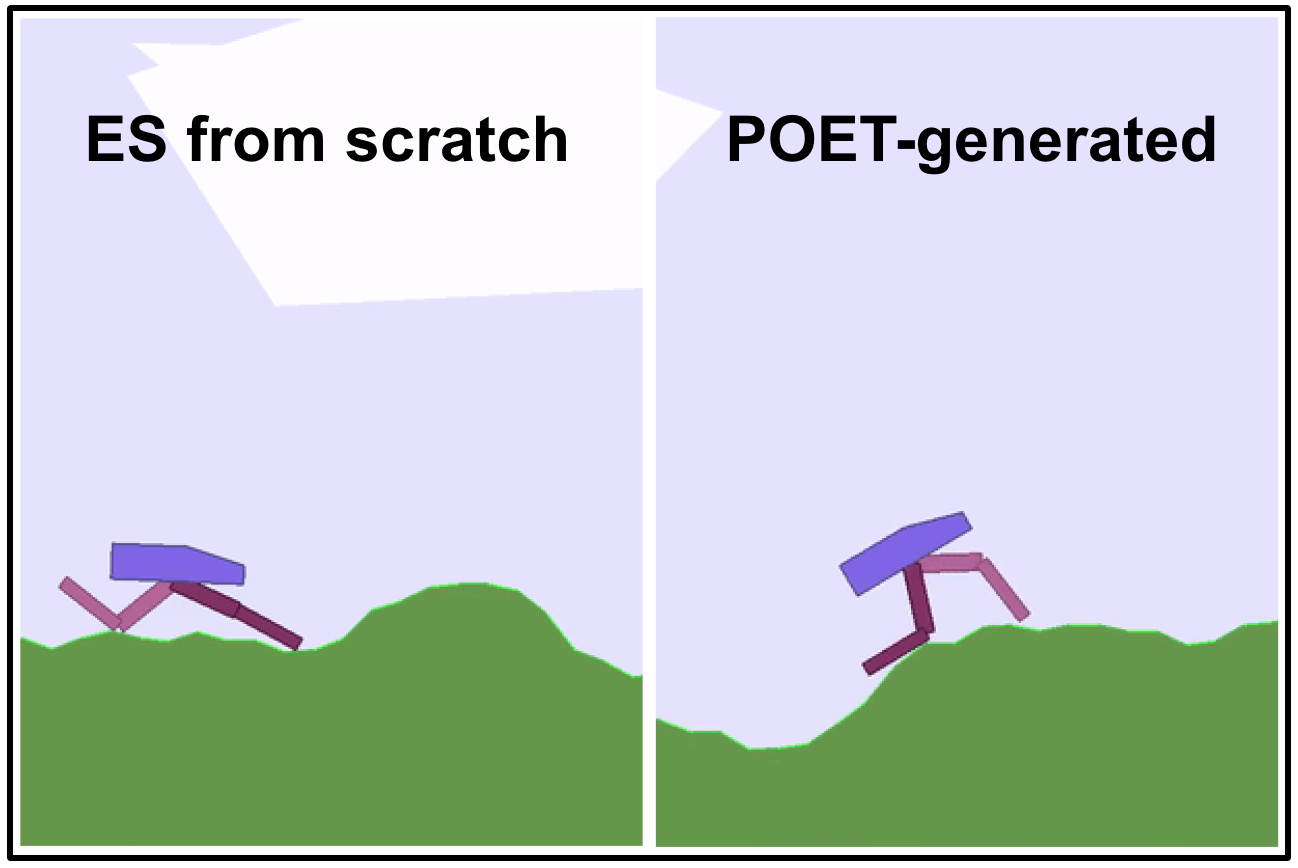}
        \caption{ES-/POET-generated agents on rough surfaces}
        \label{fig:hillside}
    \end{subfigure}
    ~ 
    \begin{subfigure}[b]{0.49\textwidth}
        \includegraphics[width=\textwidth]{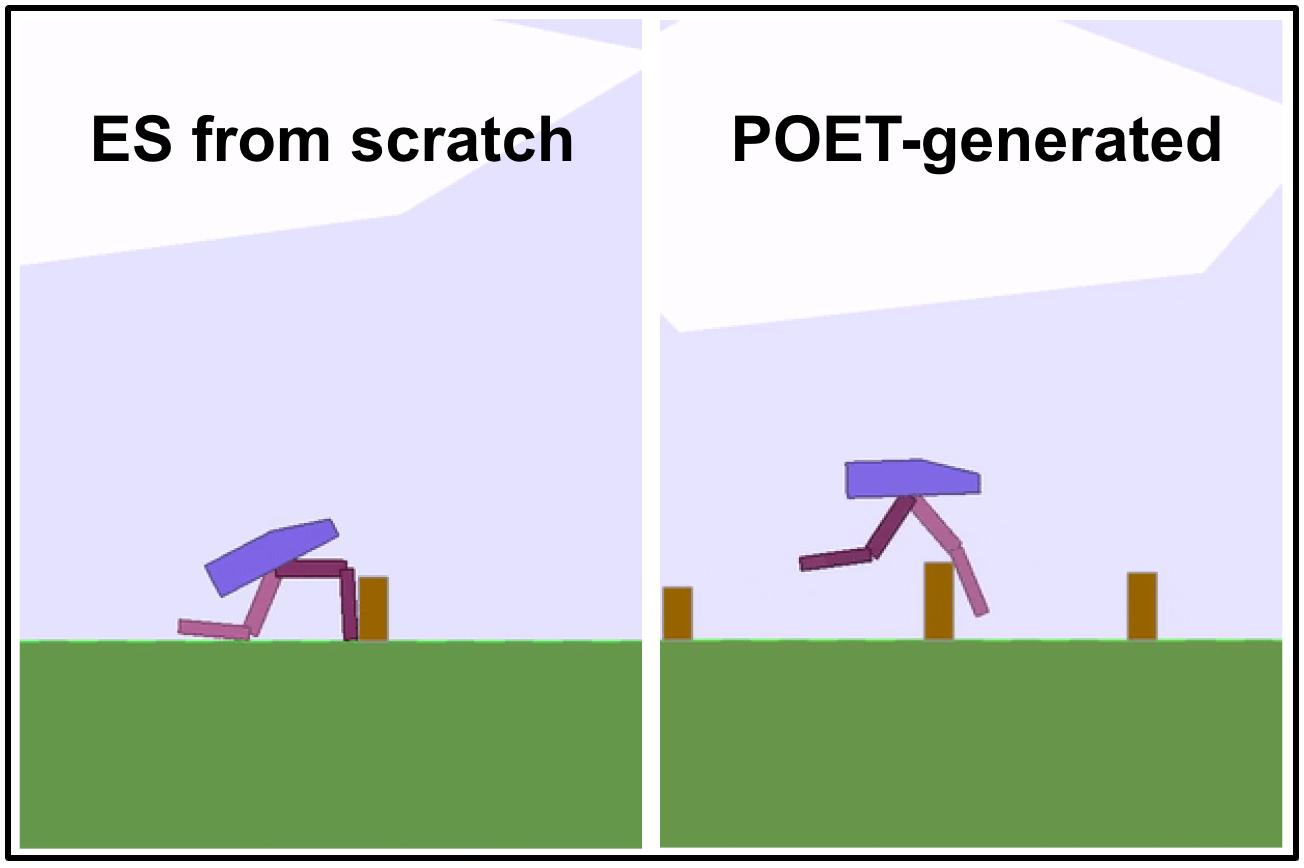}
        \caption{ES-/POET-generated agents attempting stumps}
        \label{fig:high_stump}
    \end{subfigure}
    \caption{\textbf{POET creates challenging environments and corresponding solutions that cannot be obtained by optimizing randomly initialized agents by ES.} As illustrated in the top row of (a) and the left panels of (b) and (c), agents directly optimized by ES converge on degenerate behaviors that give up early in the course. In contrast, POET not only creates these challenging environments, but also learns agents that overcome the obstacles to navigate effectively, as shown in the bottom row of (a) and right panels of (b) and (c). 
    } \label{fig:comparison_naive_control}
\end{figure}

Figure \ref{fig:comparison_naive_control} shows example results that illustrate this principle.  Three challenging environments that POET created are shown, with wide gaps, rugged hillsides, and high stumps, respectively. To demonstrate that these environments are challenging, we ran ES to directly optimize randomly initialized agents for them. More specifically, for each of the three environments, we ran ES five times with different initialization and random seeds up to 16,000 ES steps (which is twice as long as \citet{davidha:arxiv18} gives agents in the same domain with ES). Such ES-optimized agents are consistently stuck 
at local minima in these environment. 
The maximum scores out of the five ES runs for the environments illustrated in Figure \ref{fig:large_gap}, \ref{fig:hillside}, and \ref{fig:high_stump} are 17.9, 39.6, and 13.6, respectively, far below the success threshold of 230 that POET exceeds in each case.  
In effect, to obtain positive scores, these agents learn to move forward, but also to freeze before challenging obstacles, which helps them avoid the penalty of $-100$ for falling.  This behavior is a local optimum: the agents could in principle learn to 
overcome the obstacles, but instead converge on playing it safe by not moving.
Note that ES had previously been shown to be a competent approach \cite{dha_esblog} on the original Bipedal Walker Hardcore environment in OpenAI Gym, which, as illustrated later in Table \ref{challenge-table}, consists of similar, but much less difficult obstacles.
The implication is that the challenges generated and solved by POET are significantly harder and ES alone is unable to solve them.


For example, in Figure \ref{fig:large_gap}, the sequence in the top row illustrates the agent choosing to stop to avoid jumping over a dangerously wide gap: it slowly sticks one of its feet out to reach the bottom of the first wide gap and then maintains its balance without any further movement until the time limit of the episode is reached.
Videos of this and other agents from this paper can be found at \url{eng.uber.com/poet-open-ended-deep-learning}.
In contrast, POET not only \emph{creates} such challenging environments, but also in this case learns a clever behavior to overcome wide gaps and reach the finish line (second row of Figure \ref{fig:large_gap}). 

\begin{figure}
    \centering
    \begin{subfigure}[b]{0.49\textwidth}
        \includegraphics[width=\textwidth]{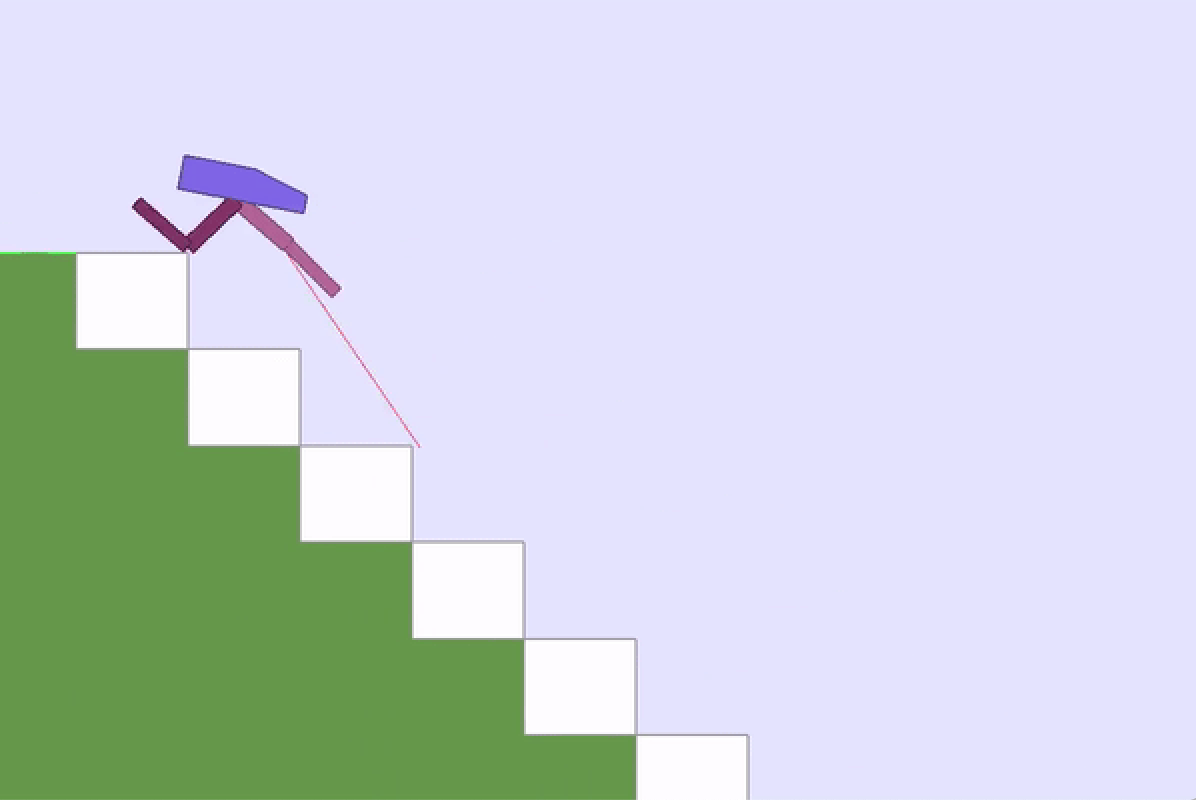}
        \caption{A downstairs course with a very large step height}
        \label{fig:downstairs_hero}
    \end{subfigure}
    ~ 
    \begin{subfigure}[b]{0.49\textwidth}
        \includegraphics[width=\textwidth]{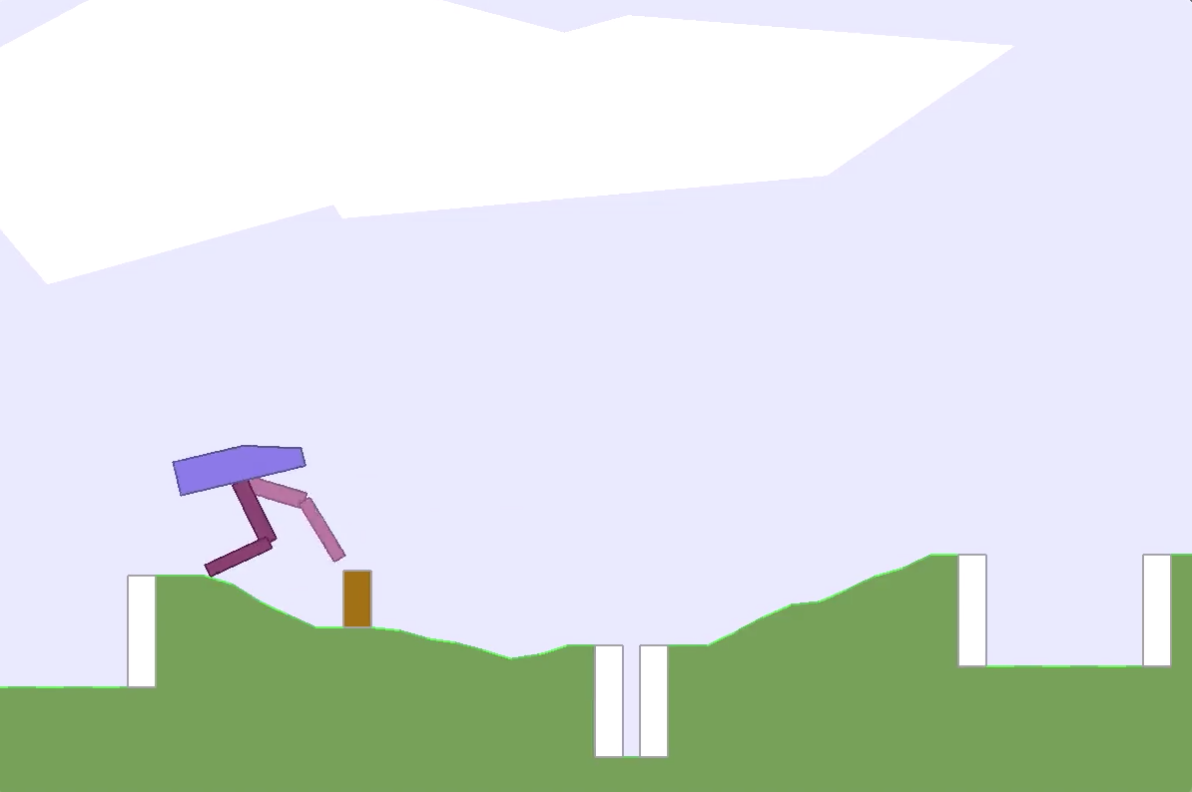}
        \caption{A rough walking course with gaps and stumps}
        \label{fig:mixed_up}
    \end{subfigure}
    \caption{\textbf{Agents exhibit specialized skills in challenging environments created by POET}.  These include the agent navigating (a) very large steps and (b) a rough walking course with mixed gaps of small and large widths, and stumps of various heights. 
    } \label{fig:challenging_envs}
\end{figure}

\begin{figure}
  \centering
  \includegraphics[width=.75\linewidth]{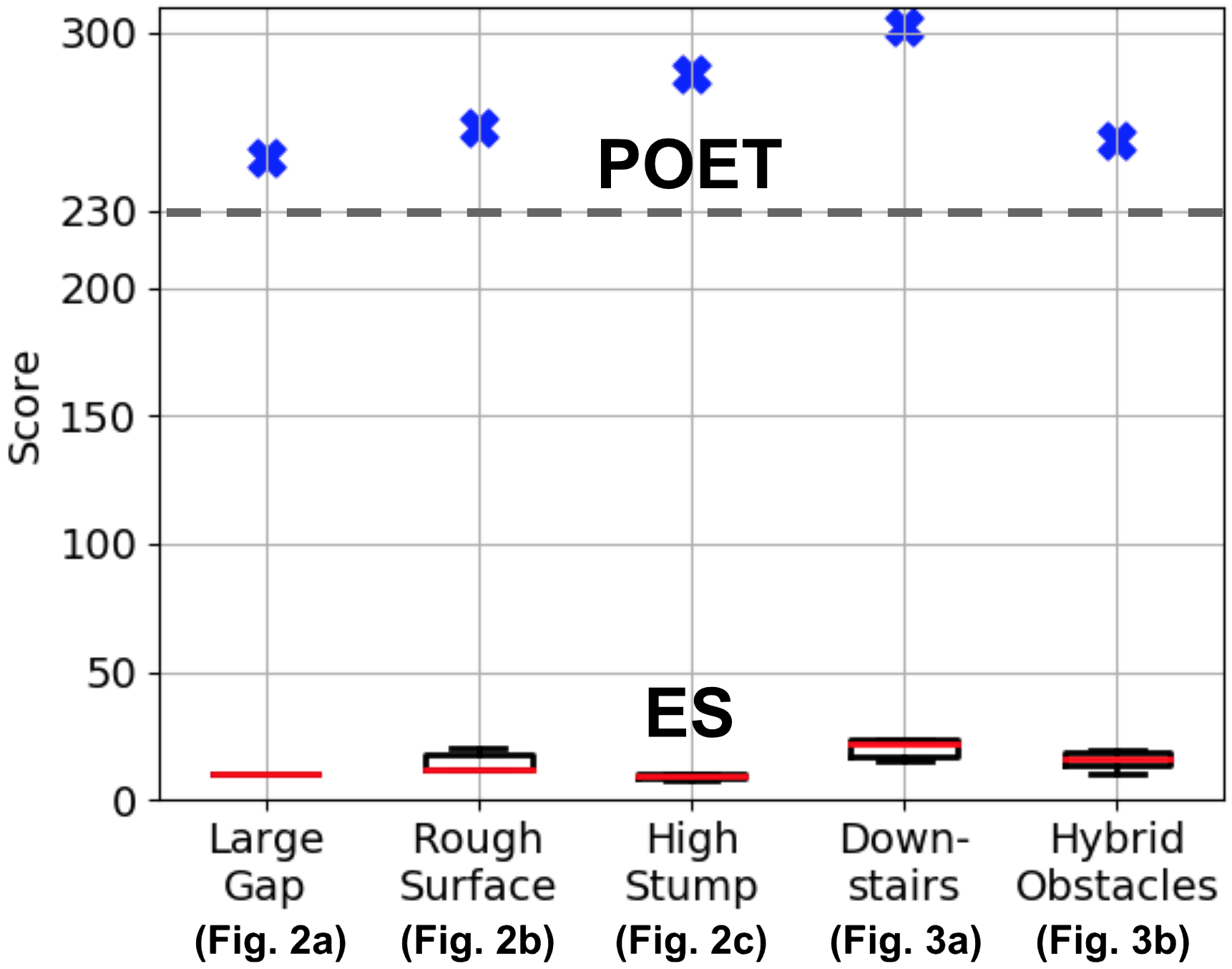}
\caption{\textbf{ES on its own cannot reproduce results achieved by POET.} Each column compares a box plot representing five ES runs to the performance achieved by POET in the single-obstacle-type problems (except for the last column) generated by POET.  Recall that a score of 230 (the dotted line) is the threshold for success.  
Though they are hard to see in these plots because of the narrow ES distributions, the red lines are medians, which are surrounded by a box from the first quartile to the third, which are further surrounded by lines touching the minimum and maximum. 
As the plot illustrates, ES on its own definitively fails in every case to come close to the challenges set and solved by POET.}
  \label{fig:compare-poet-es}
\end{figure}

Figure \ref{fig:challenging_envs} illustrates two more interesting examples of challenging environments and the highly adapted 
capabilities of their paired agents.   In the first, the agent navigates a very steep staircase of oversized steps, which ES alone cannot solve.  In the second, an experiment in heterogeneous environments (with both gaps and stumps available) yields a single obstacle course of varying gap widths and stump heights where
the POET agent succeeds and ES fails. 
The maximum scores out of the five ES runs for the environments illustrated in Figure \ref{fig:downstairs_hero} and \ref{fig:mixed_up} are 24.0 and 19.2, respectively, again far below the success theshold for POET of 230. Figure \ref{fig:compare-poet-es} illustrates the dramatic failure of ES to match the performance of POET in these relatively simple environments that POET generated and solved.

An interesting aspect of open-ended algorithms is that
statistical comparisons can be challenging 
because such algorithms are not trying to perform well on a specific preconceived target, and thus some amount of post-hoc analysis is often necessary.  That property is what
motivates an analysis driven by choosing environments produced and solved by POET, and then checking whether direct optimization can solve these environments.
However, from a statistical comparison standpoint it raises the challenge that because each environment created by POET is unique, we only have one score for POET to compare against a set of direct-optimization ES runs. For that reason, we use a \emph{single-sample t-test} that assumes the POET distribution is centered on the single POET score and measures whether the direct optimization distribution’s mean is statistically significantly lower. 
As the very poor results for ES would suggest, based on single-sample t-tests for all five environments in figures \ref{fig:comparison_naive_control} and \ref{fig:challenging_envs} that are created by POET, the scores of agents optimized by ES alone are very unlikely to be from a distribution near the POET-level solutions
($p < 0.01$). 

One interpretation of POET's ability to create agents that can solve challenging problems is that it is in effect an automatic curriculum builder. Building a proper curriculum is critical for learning to master tasks that are challenging to learn from scratch due to a lack of informative gradient. However, building an effective curriculum given a target task is itself often a major challenge. Because newer environments in POET are created through mutations of older environments and because POET only accepts new environments that are not too easy not too hard for current agents, POET implicitly builds a curriculum for learning each environment it creates.  The overall effect is that it is building many overlapping curricula simultaneously, and continually checking whether skills learned in one branch might transfer to another.

A natural question then is whether the environments created and solved by POET can also be solved by an explicit, \emph{direct-path curriculum-building control} algorithm.  To test this approach, we first collect a sample of environments generated and solved by POET, and then apply the direct-path control to each one separately to see if it can reach the same capabilities on its own. 
In this control, the agent is progressively trained on a sequence of environments of increasing difficulty that move towards the target environment. 
This kind of incremental curriculum is intuitive and variants of it appear in the literature when a task is too hard to learn directly \cite{gomez:ab97, bengio2009curriculum, karpathy2012curriculum,heess2017emergence, justesen2017procedural}.
The sequence of environments start with an environment of only flat ground without any obstacles (which is easy enough for any randomly initialized agent to quickly learn to complete). Then each of the subsequent environments are constructed by slightly modifying the current environment. More specifically, to get a new environment, each obstacle parameter of the current environment has an equal chance of staying the same value or \emph{increasing} by the corresponding mutation step value in Table \ref{env-mutation-table} until that obstacle parameter reaches that of the target environment. 

In this direct-path curriculum-building control, the agent moves from its current environment to the next one when the agent's score in the current environment reaches the reproduction eligibility threshold for POET, i.e.\ the same condition for when an environment reproduces in POET. The control algorithm optimizes the agent with ES (just as in POET). It stops when the target environment is reached and solved, or when a computational budget is exhausted. To be fair, each run of the control algorithm is given the same computational budget (measured in total number of ES steps) spent by POET to solve the environment,
which includes all the ES steps taken in the entire sequence of environments along the direct line of ancestors (taking into account transfers) leading to the target. 
This experiment is interesting because if the direct-path control cannot reach the same level as the targets, it means that a direct-path curriculum alone is not sufficient to produce the behaviors discovered by POET, supporting the importance of multiple simultaneous paths and transfers between them.

The direct-path curriculum-building control is tested against a set of environments generated and solved by POET that encompass a range of difficulties. 
For these experiments, the environments have three obstacle types enabled: gaps, roughness, and stumps.
Unlike when comparing to ES alone, he we allow POET to generate environments with multiple obstacles at the same time because the curriculum-based approach should in principle be more powerful than ES alone.
To provide a principled framework for choosing the set of generated environments to analyze, they are classified into three difficulty levels. 
The difficulty level of an environment is based on how many conditions it satisfies out of the three listed in Table \ref{challenge-table}.
In particular, a \emph{challenging} environment satisfies one of the three conditions; a \emph{very challenging} environment satisfies two of the three; and satisfying all three makes an environment \emph{extremely challenging}.
It is important to note that these conditions all merit the word ``challenging'' because they all are much more demanding than the corresponding values used in the original Hardcore version of Bipedal Walker in OpenAI Gym \cite{OpenAI_bipedal} (denoted as \emph{reference} in Table \ref{challenge-table}, which was also used by \citet{davidha:arxiv18}.  More specifically, they are 1.2, 2.0, and 4.5 times the corresponding reference values from the original OpenAI Gym environment, respectively. 

\begin{table}[H]
\begin{center}
\begin{small}
\begin{sc}
\begin{tabular}{lccccc}
\hline
 & Top Value & Top Value &  \\
 & in range of &  in range of & Roughness \\
 & Stump Height & Gap Width & \\
\hline
This Work & $\geq2.4$ & $\geq6.0$ &  $\geq4.5$\\
Reference & $2.0$ & $3.0$ & $1.0$\\
\hline
\end{tabular}
\end{sc}
\end{small}
\end{center}
\caption{\textbf{Difficulty level criteria.} The difficulty level of an environment is based on how many conditions it satisfies out of the three listed here. The \emph{reference} in the second row shows the corresponding top-of-range values used in the original Hardcore version of Bipedal Walker in OpenAI Gym \cite{OpenAI_bipedal}. The  difficulty of the environments produced and solved by POET are much higher than the reference values.}\label{challenge-table}
\vskip -0.1in
\end{table}

In this experiment, each of three runs of POET takes up to 25,200 POET iterations with a population size of 20 active environments, while the number of sample points for each ES step is 512.
These runs each take about 10 days to complete on 256 CPU cores.
The mean (with 95\% confidence intervals) POET iterations spent on solving challenging, very challenging, and extremely challenging environments starting from the iteration when they were first created are $638\pm133$, $1$,$180\pm343$, and $2$,$178\pm368$, respectively.
Here, one POET iteration refers to 
creating new environments, optimizing current paired agents, and attempting transfers (i.e.\ lines 4-22 in Algorithm \ref{alg:poet}). 
The more challenging environments clearly take more effort to solve.

\begin{figure}
  \centering
  \includegraphics[width=.83\linewidth]{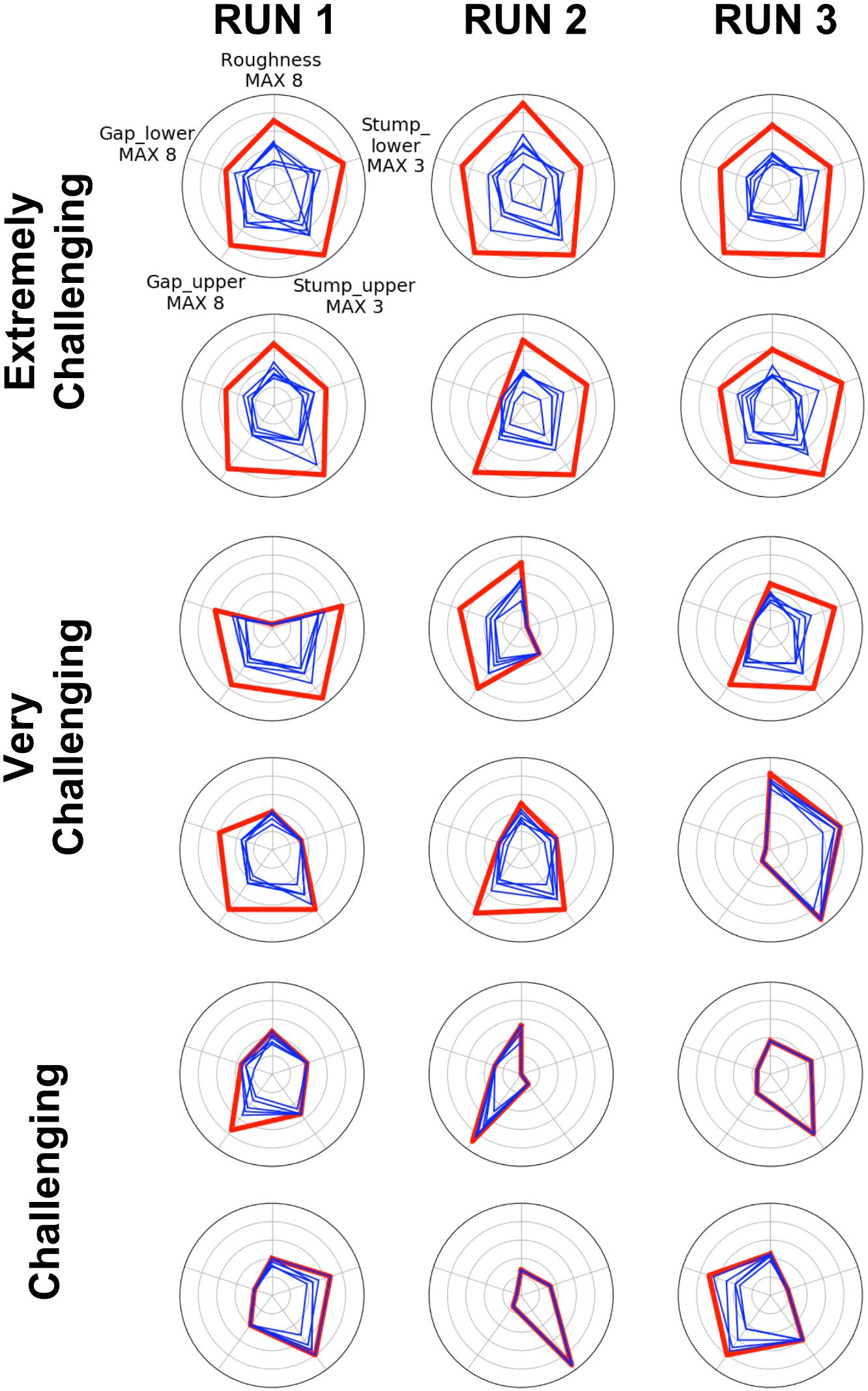}
\caption{\textbf{POET versus direct-path curriculum-building controls.} Each rose plot depicts one environment that POET created and solved (red pentagon). For each, the five blue pentagons indicate what happens in control runs when the red pentagon is the target. 
Each blue pentagon is the closest-to-target environment solved by one of the five independent runs of the control algorithm. The five vertices of each pentagon indicate roughness (\emph{Roughness}), the bottom and top values of the range of the gap width of all the gaps (\emph{Gap\_lower} and \emph{Gap\_upper}), and the bottom and top values for the height of stumps  (\emph{Stump\_lower} and \emph{Stump\_upper}) in the given solved environment. The value after MAX in the key is the maximum value at the outermost circle for each type of obstacle. Each column contains sample solved environments from a single independent run of POET.
}
  \label{fig:control_type_1}
\end{figure}

Figure \ref{fig:control_type_1} compares the POET environments and the direct-path curriculum-building control algorithm through a series of \emph{rose plots}, wherein each rose plot compares the configuration of an environment solved by POET (red pentagons) with the closest that the direct-path control could come to that configuration (blue pentagons).  For each red pentagon there are five such blue pentagons, each representing one of five separate attempts by the control to achieve the red pentagon target.
The five vertices of each pentagon indicate roughness, the lower and upper bounds of the range of the gap width, and those of the stump height, respectively.
Each column in figure \ref{fig:control_type_1} consists of six representative samples (the red pentagons) of environments that a single run of POET up to 25,200 iterations 
created and solved.  
As the rows descend from top to bottom, the difficulty level decreases.  In particular, the top two rows are extremely challenging targets, the next two are very challenging, and the bottom two are challenging (all are randomly sampled from targets generated and solved at each difficulty level in each run).  
Qualitatively, it is clear from figure \ref{fig:control_type_1} that attempts by the direct-path control consistently fail to reach the same level as POET-generated (and solved) levels.  


To more precisely quantify these results, we define the normalized distance between any two environments, $E_{A}$ and $E_{B}$ as
$ \frac{1}{\beta}\vert\vert{\frac{e(E_{A}) - e( E_{B})}{e(E_{\textrm{Max}})}}\vert\vert_{2}$, where $e(E)$ is the genetic encoding vector of $E$, and $E_{\textrm{Max}}$ is a hypothetical environment (for normalization purpose) with all genetic encoding values are maxed out. (The maximum values are $\textrm{roughness} = 8, \textrm{Gap\_lower}=\textrm{Gap\_upper} = 8, \textrm{Stump\_lower} = \textrm{Stump\_upper} = 3$.)
The constant $\beta = \sqrt{5}$ is simply for normalization so that the distance is normalized to $1$ when $E_{A} = E_{\textrm{Max}}$ and $E_{B}$ is a purely flat environment (roughness zero) without any obstacles, which therefore contains an all-zero genetic encoding vector. 
Based on this distance measure, we can calculate the median values and confidence intervals of distances between target environments (created and solved by POET) and the corresponding closest-to-target environments that the control algorithms can solve. 
These values are calculated for all the environments of different challenge levels across all three runs shown in figure \ref{fig:control_type_1}. 

\begin{figure}
  \centering
  \includegraphics[width=.85\linewidth]{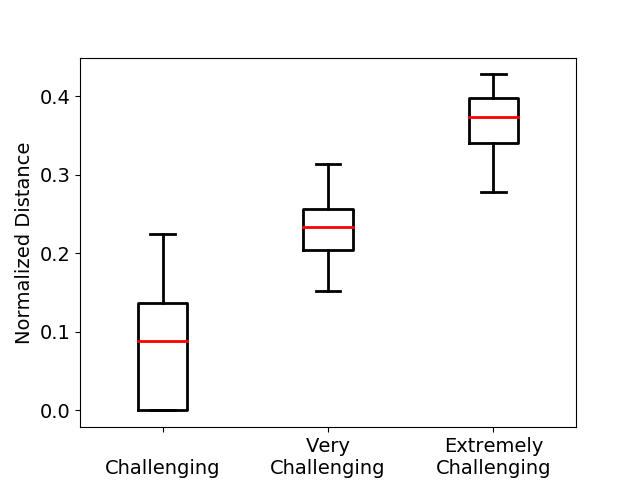}
\caption{\textbf{Distances between targets and results of direct-path curriculum-building control runs.} Here, levels \emph{challenging}, \emph{very challenging}, and \emph{extremely challenging} correspond to the bottom two, middle two and top two rows of red pentagons in Figure \ref{fig:control_type_1}, respectively.
The red lines are medians, which are surrounded by a box from the first quartile to the third, which are further surrounded by lines touching the minimum and maximum.
These results highlight how difficult it becomes for the direct-path control to match challenges invented and solved by POET as the challenge level increases.
}
  \label{fig:distance-boxplot}
\end{figure}

The results, summarized in figure \ref{fig:distance-boxplot}, demonstrate that POET creates and solves environments that the control algorithm fails to solve at very and extremely challenging difficulty levels, while the curriculum-based control algorithm can often (though not always) solve environments at the lowest challenge level. 
One implication of these results is that the direct-path curriculum-building control is valid in the sense that it does perform reasonably well at solving minimally challenging scenarios.  However, the very and extremely challenging environments that POET invents reach significantly beyond what the direct-path curriculum can match.
There are a couple ways to characterize this significance level more formally.   These analyses are based on the \emph{distance between the closest environment reached by the control and the POET-generated target}.  First, we can examine whether that distance is significantly greater the higher the challenge level.  Indeed, Mann-Whitney U tests 
show that the difference between such distances between challenging and very challenging environments is indeed very significant ($p < 0.01$), as is the difference between such distances between very challenging and extremely challenging.  This test gives a quantitative confirmation of the intuition (also supported by figure \ref{fig:distance-boxplot}) that these challenge levels are indeed meaningful, and the higher they go, the more out of reach they become for the control.  Second, 
a single-sample t-test (which was previously used when comparing POET to the ES-alone control)
can also here provide confidence that the distances from the targets are indeed far in an absolute sense: indeed, even at the lowest challenge level (and for all challenge levels), the one-sample t-test measures high significance ($p < 0.01$) between the distribution of distances from the target and the target level.


In aggregate, the results from the direct-path curriculum-building control
help to show quantitatively the advantage of POET over conventional curriculum-building. In effect, the ability to follow multiple chains of environments in the same run and transfer skills among them pushes the frontier of skills farther than a single-chain curriculum can push, hinting at the limitations of preconceived curricula in general. 
Thus, while POET cannot guarantee reaching a particular preconceived target,
these results suggest that in some environment spaces POET's unique ability to generate multiple different challenges and solve them may actually still provide a more promising path towards solving some preconceived target challenges (i.e. offering a powerful alternative to more conventional optimization), putting aside the additional interesting benefits of an algorithm that invents, explores, and solves new types of challenges automatically.

A fundamental problem of a pre-conceived direct-path curriculum (like the control algorithm above) 
is the potential lack of necessary \emph{stepping stones}. In particular, skills learned in one environment can be useful and critical for learning in another environment. Because there is no way to predict where and when stepping stones emerge, the need arises to conduct transfer experiments (which POET implements) from differing environments or problems \cite{nguyen2016innovative}. In conventional curriculum-building algorithms, the ``transfer'' only happens once, i.e.\ at the time when the new environment is created, and is unidirectional, i.e.\ from the agent paired with the parent environment (\emph{parent agent}) to the \emph{child environment} with the hope that the parent agent helps jump-start learning in the child environment. 

\begin{figure}
  \centering
     \begin{subfigure}[b]{0.9\textwidth}
        \includegraphics[width=\textwidth]{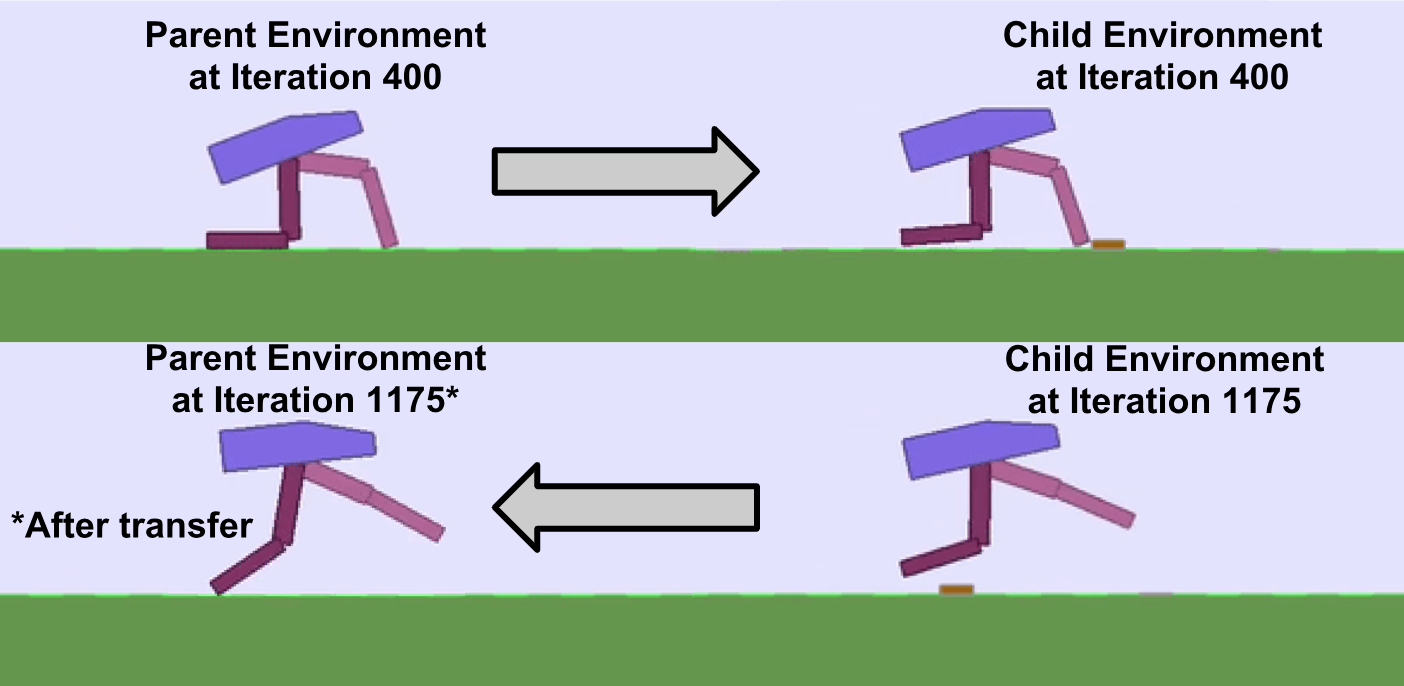}
        \caption{Transfer from agent in parent environment to child environment and vice versa}
        \label{fig:transfer_2stage}
    \end{subfigure}
    ~ 
    \begin{subfigure}[b]{0.45\textwidth}
        \includegraphics[width=\textwidth]{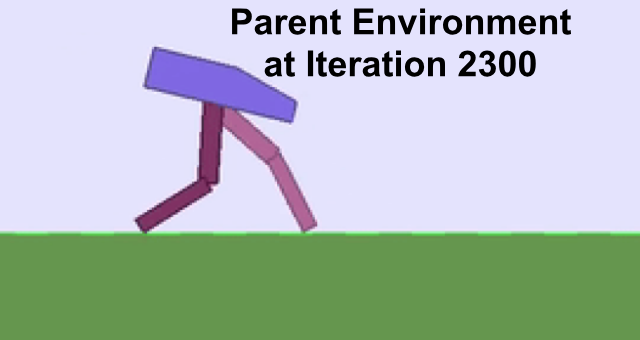}
        \caption{The walking gait of agent in parent environment at Iteration 2,300}
        \label{fig:transfer_before_after}
    \end{subfigure}
\caption{\textbf{Synergistic two-way transfer between parent and child environments.} At iteration 400, a transfer from parent environment yields a child agent now learning in a stumpy environment, shown in the top row of (a). The child agent eventually learns to stand up and jump over the stumps, and at iteration 1,175 that skill is transferred \emph{back} to the parent environment, depicted in the bottom row of (a). This transfer from the child environment back to the parent helps the parent agent learn a more optimal walking gait compared to its original one. Given the same amount of computation, the agent with the original walking gaits reaches a score of 309, while the one with the more optimal walking gait as illustrated in (b) reaches a score of 349.
}
  \label{fig:transfer}
\end{figure}

In contrast, POET maintains parallel paths of environment evolution and conducts transfer experiments periodically that give every active agent multiple chances to transfer its learned skills to others, including the transfer of a child agent back to its parent environment, which continually creates and preserves opportunities to harness unanticipated stepping stones. Figure \ref{fig:transfer} presents an interesting example from POET where such transfers help a parent environment acquire a better solution: The parent environment includes only flat ground and the parent agent learns to move forward without fully standing up (top left), which works in this simple environment, but represents a local optimum in walking behavior because standing would be better. At iteration 400, the parent environment mutates and generates a child environment with small stumps, and the child agent inherits the walking gait from the parent agent. 

At first, it can move forward in the new stumpy environment, but it often stumbles because of its crouching posture (top right). Later, through several hundred further iterations of ES, the child agent eventually learns to stand up and jump over the stumps.
The interesting moment is iteration 1,175, when that skill is transferred back to the \emph{parent} environment (middle row).  For the first time, the parent environment (which is completely flat) now contains an agent who stands up while walking.   Without the backward transfer, the agent in the simple flat environment would be stuck at its original gait (which is in effect a local optimum) forever, which was confirmed by separately turning off transfer and running the kneeling (local optimum) agent in the original (parent) environment for 3,000 iterations, which did not result in any substantive change.
Interestingly, after transfer back to the parent environment, the transferred agent then continued to optimize the standing gait to the parent (stump-free) environment and eventually obtained a much better walking gait that moves faster and costs less energy due to less friction (bottom row). More specifically, given 3,000 iterations, the agent with the original kneeling walking gait
reaches a score of 309, while the one with the more optimal walking gait illustrated in figure \ref{fig:transfer_before_after} reaches a score of 349.

To give a holistic view of the success of transfer throughout the entire system, we count the number of \emph{replacement attempts} during the course of a run, which means the number of times an environment took a group of incoming transfer attempts from all the other active environments.
The total number of such replacement attempts in RUN 1, RUN 2, and RUN 3 (labelled in Figure \ref{fig:control_type_1}) are 18,894, 19,014, and 18,798, respectively, out of which, $53.62\%$, $49.26\%$, $48.89\%$, respectively, are successful replacements (meaning one of the tested agents is better than the current niche champion). 
Note that each replacement attempt here encompasses both the direct transfer and proposal transfer attempts. 
These statistics show how pervasively transfer permeates (and often adds value to) the parallel paths explored by POET.

\begin{figure}
  \centering
  \includegraphics[width=.85\linewidth]{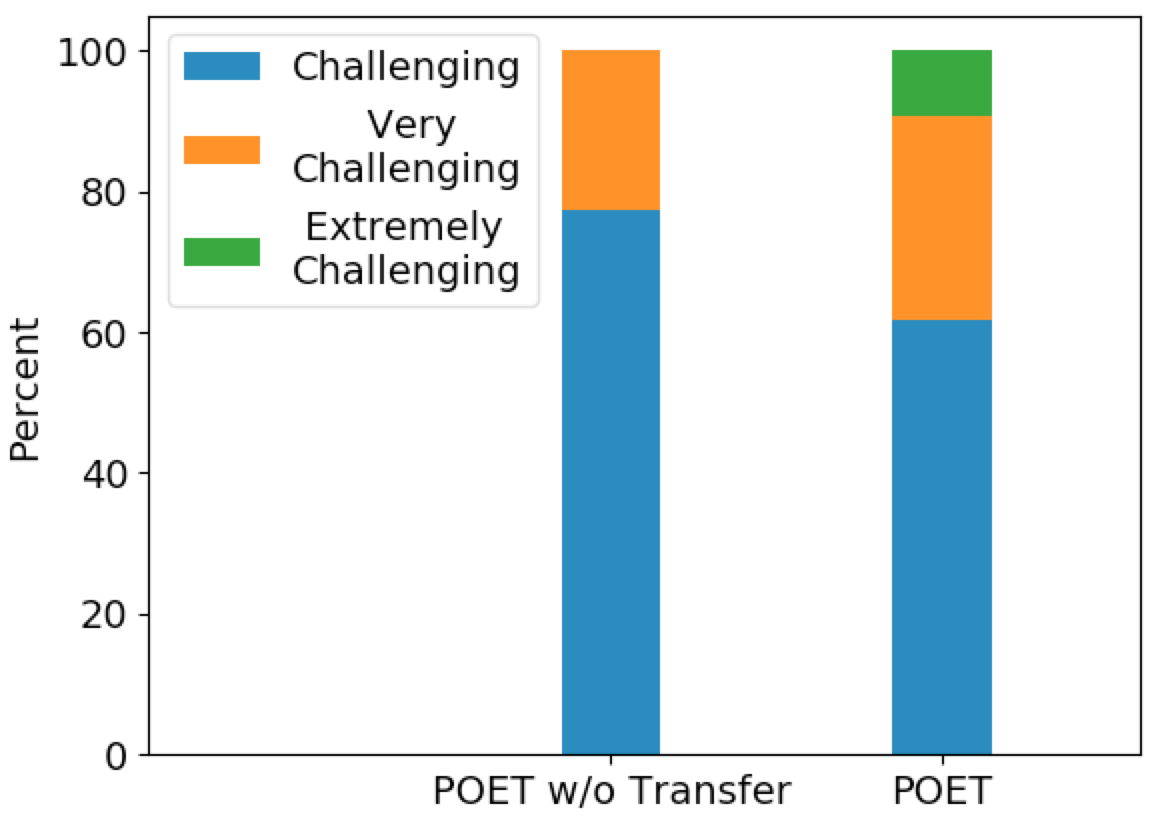}
\caption{\textbf{Percentage of environments of different difficult levels created and solved by POET and the control runs of POET without transfer.} In the control without transfer, no \emph{extremely challenging} environments are generated.
}
  \label{fig:disable-transfer-barplot}
\end{figure}

While transfer is pervasive, that does not in itself prove it is essential. 
To demonstrate the value of transfer, a control is needed: We relaunched another three POET runs, but with all the transfers \emph{disabled} (which we call \emph{POET without transfer}). In this variant, POET runs as usual, but simply never
tries to transfer paired solutions from one environment to another.
We can then calculate the \emph{coverage} of the environments that are created and solved by POET and the control, respectively, following a similar metric as defined in \citet{lehman:gecco11}: We first uniformly sample 1,000 \emph{challenging}, 1,000 \emph{very challenging}, and 1,000 \emph{extremely challenging} environments. For each of the total 3,000 sampled environments, the distance to the nearest
one of the \emph{challenging}, \emph{very challenging}, or \emph{extremely challenging} environments created and solved by POET (and by the control, respectively) is calculated. Note that the better covered the environment space is, the lower the sum of all such nearest distances will be. The result is that the coverage of environments created and solved by POET is significantly greater than by POET without transfer ($p < 2.2e{-16}$ based on Mann-Whitney U test), meaning POET with transfer explores more of the space of environments, including creating (and solving) more difficult environments. 
In an even more dramatic statistic showing the essential role of transfer in open-ended search, in POET without transfer, \emph{no extremely challenging environments are solved at all (Fig. \ref{fig:disable-transfer-barplot}).
}

\begin{figure}
  \centering
  \includegraphics[width=.8\linewidth]{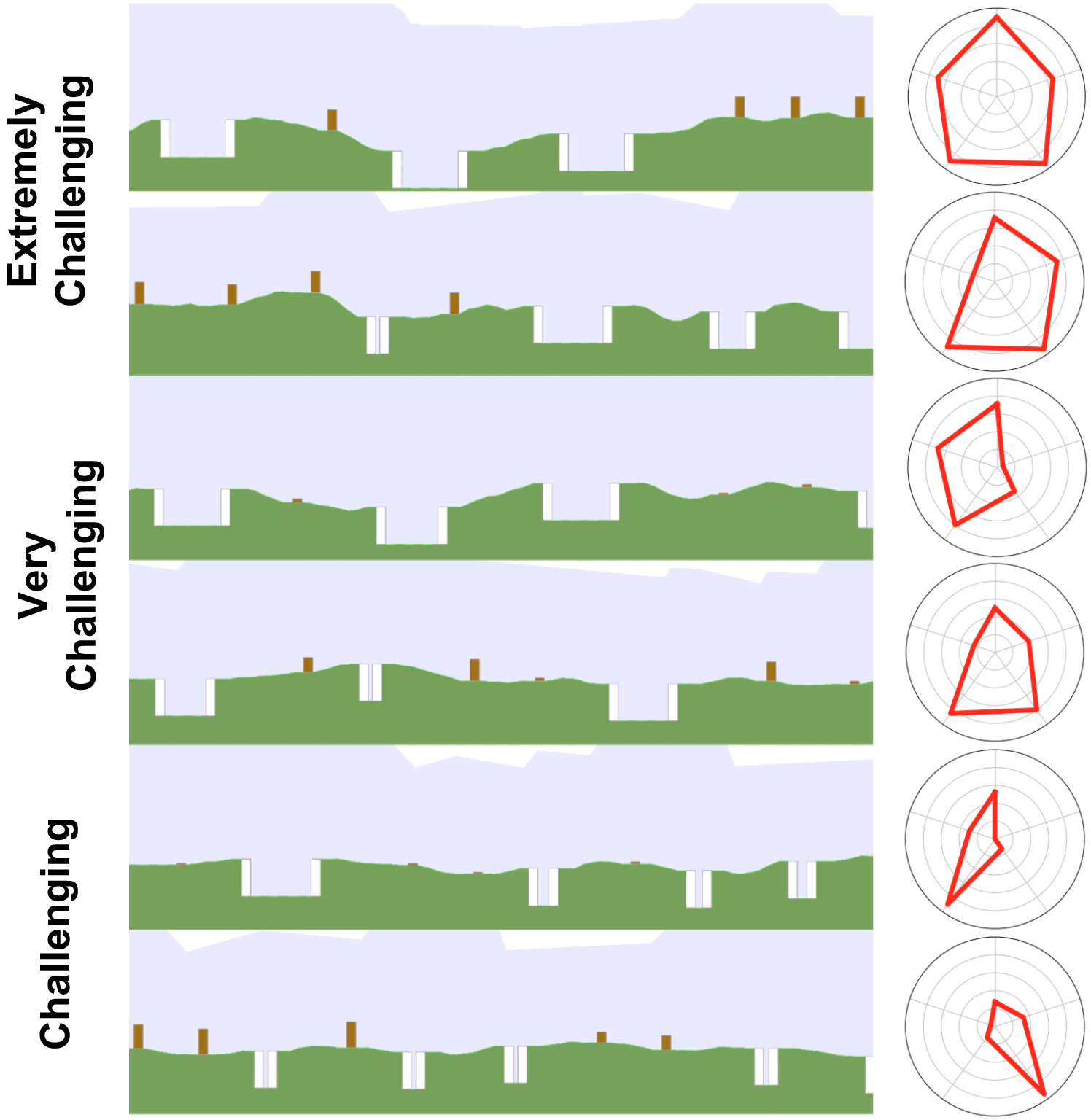}
\caption{\textbf{Sub-sections of environments from the middle column (RUN 2) of figure \ref{fig:control_type_1}.} These environments exhibit diversity in the levels and ranges of different obstacles. For example, the topmost environment has all large gaps and high stumps with a high amount of roughness; the second from the bottom exhibits a wide variety in gap width, but with trivial stumps and low roughness. In contract, the environment at the very bottom exhibits a wide range of stump heights, but with low gap width and low roughness.
}
  \label{fig:diverse_envs}
\end{figure}

Finally, one of the primary hypothesized benefits of POET is its ability to produce a broad diversity of different problems with functional solutions in a single run. The environments (depicted as red pentagons) shown in each column of Figure \ref{fig:control_type_1} are created and solved in a \emph{single} run of POET. 
Each such column exhibits  diversity in the values and/or value ranges in roughness, gap width of gaps, and height of stumps. Take the environments in the middle column (labelled ``RUN 2'') for example. In figure \ref{fig:diverse_envs},
each image of a section of an environment from top to bottom corresponds to the same row of rose plot for RUN 2 in Figure \ref{fig:control_type_1}.
For instance, the topmost environment has narrow ranges and high values in gap width and stump height, and high value in roughness; in stark contrast the plot at the bottom depicts a less challenging environment with a wide range of stump heights, but low gap width and roughness.  
This diversity of environments also implies diverse experiences for the agents paired with them, who in turn thereby learn diverse walking gaits.  This diversity then supplies the stepping stones that fuel the mutual transfer mechanism of POET.
For a more objective statement of POET's ability to cover much of the possibility space through its diversity, all three POET runs created and solved sufficiently diverse environments to cover all three challenge levels.

\section{Discussion, Future Work, and Conclusion}

The promise of open-ended computation is fascinating for its potential to enable systems that become more powerful and interesting the longer they run.  There is always the possibility that the feats we observe in the system today will be overshadowed by the achievements of tomorrow. 
In these unfolding odysseys there is an echo of the natural world.  More than just the story of a single intelligent lifetime, they evoke the history of invention, or of natural evolution over the eons of Earth.  These are processes that produce not just a single positive result, but an ongoing cacophony of surprises -- unplanned advances rolling ahead in parallel without any final destination.  

POET is an attempt to move further down the road towards these kinds of systems.  While the road remains long, the rewards for machine learning of beginning to capture the character of open-ended processes is potentially high.  First, as the results show, there is the opportunity to discover capabilities that could not be learned in any other way, even through a carefully crafted curriculum targeted at the desired result.  In addition, a diversity of such results can be generated in a single run, and the problems and solutions can both increase in complexity over time.  Furthermore, the implicit result is that POET is self-generating multiple curricula simultaneously, all while leveraging the results of some as stepping stones to progress in others.   

The experiment in this article in 2-D walking establishes the potential of POET, but POET becomes more interesting the more unbounded its problem space becomes.  The present problem space is a 2-D course of obstacles that can be generated within distributions defined by the genome describing the environment.  This space is limited by the maximal
ranges of those distributions.  For example, there is a maximum possible gap width and stump height, which means that the system in effect can eventually ``max out'' the difficulty.  While sufficiently broad to demonstrate POET's functionality, in the future much more flexible or even unbounded encodings can enable POET to traverse a far richer problem space.  For example, an indirect encoding like a compositional pattern-producing network (CPPN) \cite{stanley:gpem07} can generate arbitrarily complex patterns that can be e.g.\ converted into levels or obstacle courses.  Such an encoding would allow POET to diverge across a much richer landscape of possibilities.

An additional constraint that exists in this initial work is that the body of the agent is fixed, ultimately limiting the sort of obstacles it can overcome (e.g.\ how big of a gap it can jump over). Here too a more powerful and expressive encoding of morphologies, including those like CPPNs that are based on developmental biology, could allow us to co-evolve the morphology and body of the agent along with its brain in addition to the environment it is solving. Previous research has shown that using such indirect encodings to evolve morphologies can improve performance and create a wide diversity of interesting solutions \cite{cheney:gecco13}. Even within the 2-D Bipedal Walker domain from this paper, research has shown that allowing the morphology to be optimized can improve performance and provide a diversity of interesting solutions \cite{davidha:arxiv18}.

Furthermore, the  CPPN indirect encoding can also encode the neural network controllers in an algorithm called HyperNEAT \cite{stanley:alife09}.  Work evolving robot controllers with HyperNEAT has shown it can produce neural networks with regular geometric patterns in neural connectivity that exploit regularities in the environment to produce regular behaviors that look natural and improve performance \cite{clune:ieeetec11, clune:cec09, clune:cec09, yosinski:ecal11}. POET allows us to create the interesting combination of using an indirect encoding like a CPPN to encode the environment, the body of the agent, and the neural network controller of that agent, allowing each of the three to benefit from and exploit the regularities being produced by the others.

While ES is the base optimization algorithm in this paper for training solvers for various tasks under POET, including the main transfer mechanism, POET can be instantiated with any RL or optimization algorithm.   Trust Region Policy Optimization (TRPO) \cite{schulman2015trust}, Proximal Policy Optimization (PPO) \cite{ppo:arxiv17}, genetic algorithms, other variants of ES, and many other such algorithms are all viable alternatives.
This plug-and-play aspect of the overall framework presents an intriguing set of future opportunities to hybridize the divergent search across environments with the most appropriate inner-loop learning algorithms, opening up the investigation of open-endedness itself to diverse research communities.

POET could also substantially drive progress in the field of meta-learning, wherein neural networks are exposed to many different problems and get better over time at learning how to solve new challenges (i.e.\ they learn to learn). Meta-learning requires access to a distribution of different tasks, and that traditionally requires a human to specify this task distribution, which is costly and may not be the right or best distribution on which to learn to learn. Gupta et al. \cite{gupta2018unsupervised} note both that the performance of meta-learning algorithms critically depends on the distribution of tasks they meta-train on, and that automatically producing an effective distribution of tasks for meta-learning algorithms would represent a substantial advance. That work took an important initial step in that direction by automatically creating different reward functions within one fixed environment \cite{gupta2018unsupervised}. POET takes the important further step of creating entirely new environments (or challenges, more abstractly). While the version of POET in this initial work does not also also create a unique reward function for each environment, doing so is a natural extension that we leave for future work. One possibility is that the reward function could be encoded in addition to the rest of the environment and optimized similarly in parallel.  

Another promising avenue for exploration in POET is its interaction with generalization versus specialization.  Interestingly, the environments in POET themselves can foster generalization by including multiple challenges in the same sequence.  However, alternate versions of POET are also conceivable that take a more explicit approach to optimizing for generality; for example, compound environments could be created whose scores are the aggregate of scores in their constituent environments (borrowing a further aspect of the CMOEA algorithm \cite{huizinga2016does, Joost:arxiv18}).  

Finally, it is exciting to consider for the future the rich potential for surprise in all the possible domains where POET might be applied.  For example, 3-D parkour was explored by \citet{heess:arxiv17} in environments created by humans, but POET could invent its own creative parkour challenges and their solutions.  The soft robots evolved by \citet{cheney:gecco13} would also be fascinating to combine with ever-unfolding new obstacle courses.  POET also offers practical opportunities in domains like autonomous driving, where through generating increasingly challenging and diverse scenarios it could uncover important edge cases and policies to solve them.  Perhaps more exotic opportunities could be conceived, such as protein folding for specific biological challenges invented by the system, or searching for new kinds of chemical processes that solve unique problems.  The scope is broad for imagination and creativity in the application of POET.  

\section*{Acknowledgments}

We thank all of the members of Uber AI Labs, in particular Joost Huizinga and Zoubin Ghahramani for helpful discussions.
We also thank Alex Gajewski (from his internship at Uber AI Labs) for help with implementation and useful discussions.   
Finally, we thank Justin
Pinkul, Mike Deats, Cody Yancey, Joel Snow, Leon Rosenshein and the entire OpusStack Team inside Uber for providing our computing platform and for technical support.

\bibliographystyle{unsrtnat}{\small\bibliography{nips,ucf,nn,nnstrings}}

\begin{thebibliography}{82}
\providecommand{\natexlab}[1]{#1}
\providecommand{\url}[1]{\texttt{#1}}
\expandafter\ifx\csname urlstyle\endcsname\relax
  \providecommand{\doi}[1]{doi: #1}\else
  \providecommand{\doi}{doi: \begingroup \urlstyle{rm}\Url}\fi

\bibitem[Deng et~al.(2009)Deng, Dong, Socher, Li, Li, and
  Fei-Fei]{Deng2009ImageNetAL}
Jia Deng, Wei Dong, Richard Socher, Li-Jia Li, Kai Li, and Li~Fei-Fei.
\newblock Imagenet: A large-scale hierarchical image database.
\newblock \emph{2009 IEEE Conference on Computer Vision and Pattern
  Recognition}, pages 248--255, 2009.

\bibitem[He et~al.(2015{\natexlab{a}})He, Zhang, Ren, and Sun]{he:arxiv15}
Kaiming He, Xiangyu Zhang, Shaoqing Ren, and Jian Sun.
\newblock Deep residual learning for image recognition.
\newblock \emph{arXiv preprint arXiv:1512.03385}, 2015{\natexlab{a}}.

\bibitem[Russakovsky et~al.(2014)Russakovsky, Deng, Su, Krause, Satheesh, Ma,
  Huang, Karpathy, Khosla, and et~al.]{russakovsky:arxiv14}
O.~Russakovsky, J.~Deng, H.~Su, J.~Krause, S.~Satheesh, S.~Ma, Z.~Huang,
  A.~Karpathy, A.~Khosla, and M.~Bernstein et~al.
\newblock Imagenet large scale visual recognition challenge.
\newblock \emph{arXiv preprint arXiv:1409.0575}, 2014.

\bibitem[He et~al.(2015{\natexlab{b}})He, Zhang, Ren, and Sun]{he2:arxiv15}
Kaiming He, Xiangyu Zhang, Shaoqing Ren, and Jian Sun.
\newblock Delving deep into rectifiers: Surpassing human-level performance on
  imagenet classification.
\newblock \emph{arXiv preprint arXiv:1502.01852}, 2015{\natexlab{b}}.

\bibitem[Sutton and Barto(1998)]{sutton1998reinforcement}
Richard~S Sutton and Andrew~G Barto.
\newblock \emph{Reinforcement learning: An introduction}, volume~1.
\newblock 1998.

\bibitem[Bellemare et~al.(2013)Bellemare, Naddaf, Veness, and
  Bowling]{bellemare2013arcade}
Marc~G Bellemare, Yavar Naddaf, Joel Veness, and Michael Bowling.
\newblock The arcade learning environment: An evaluation platform for general
  agents.
\newblock \emph{J. Artif. Intell. Res.(JAIR)}, 47:\penalty0 253--279, 2013.

\bibitem[Mnih et~al.(2015)Mnih, Kavukcuoglu, Silver, Rusu, Veness, Bellemare,
  Graves, Riedmiller, Fidjeland, Ostrovski, et~al.]{mnih2015}
Volodymyr Mnih, Koray Kavukcuoglu, David Silver, Andrei~A Rusu, Joel Veness,
  Marc~G Bellemare, Alex Graves, Martin Riedmiller, Andreas~K Fidjeland, Georg
  Ostrovski, et~al.
\newblock Human-level control through deep reinforcement learning.
\newblock \emph{Nature}, 518\penalty0 (7540):\penalty0 529--533, 2015.

\bibitem[Espeholt et~al.(2018)Espeholt, Soyer, Munos, Simonyan, Mnih, Ward,
  Doron, Firoiu, Harley, Dunning, et~al.]{espeholt2018impala}
Lasse Espeholt, Hubert Soyer, Remi Munos, Karen Simonyan, Volodymir Mnih, Tom
  Ward, Yotam Doron, Vlad Firoiu, Tim Harley, Iain Dunning, et~al.
\newblock Impala: Scalable distributed deep-rl with importance weighted
  actor-learner architectures.
\newblock \emph{arXiv preprint arXiv:1802.01561}, 2018.

\bibitem[Horgan et~al.(2018)Horgan, Quan, Budden, Barth-Maron, Hessel,
  Van~Hasselt, and Silver]{horgan2018distributed}
Dan Horgan, John Quan, David Budden, Gabriel Barth-Maron, Matteo Hessel, Hado
  Van~Hasselt, and David Silver.
\newblock Distributed prioritized experience replay.
\newblock \emph{arXiv preprint arXiv:1803.00933}, 2018.

\bibitem[Hessel et~al.(2018)Hessel, Soyer, Espeholt, Czarnecki, Schmitt, and
  van Hasselt]{hessel2018multi}
Matteo Hessel, Hubert Soyer, Lasse Espeholt, Wojciech Czarnecki, Simon Schmitt,
  and Hado van Hasselt.
\newblock Multi-task deep reinforcement learning with popart.
\newblock \emph{arXiv preprint arXiv:1809.04474}, 2018.

\bibitem[Kapturowski et~al.(2019)Kapturowski, Ostrovski, Dabney, Quan, and
  Munos]{kapturowski2018recurrent}
Steven Kapturowski, Georg Ostrovski, Will Dabney, John Quan, and Remi Munos.
\newblock Recurrent experience replay in distributed reinforcement learning.
\newblock In \emph{International Conference on Learning Representations}, 2019.
\newblock URL \url{https://openreview.net/forum?id=r1lyTjAqYX}.

\bibitem[Ecoffet et~al.(2018)Ecoffet, Huizinga, Lehman, Stanley, and
  Clune]{goexplore_blog}
Adrien Ecoffet, Joost Huizinga, Joel Lehman, Kenneth~O. Stanley, and Jeff
  Clune.
\newblock Montezuma’s revenge solved by go-explore, a new algorithm for
  hard-exploration problems (sets records on pitfall, too).
\newblock \url{https://eng.uber.com/go-explore/}, 2018.

\bibitem[Silver et~al.(2016)Silver, Huang, Maddison, Guez, Sifre, Van
  Den~Driessche, Schrittwieser, Antonoglou, Panneershelvam, Lanctot,
  et~al.]{SilverHuangEtAl16nature}
David Silver, Aja Huang, Chris~J Maddison, Arthur Guez, Laurent Sifre, George
  Van Den~Driessche, Julian Schrittwieser, Ioannis Antonoglou, Veda
  Panneershelvam, Marc Lanctot, et~al.
\newblock Mastering the game of go with deep neural networks and tree search.
\newblock \emph{nature}, 529\penalty0 (7587):\penalty0 484, 2016.

\bibitem[Silver et~al.(2017)Silver, Schrittwieser, Simonyan, Antonoglou, Huang,
  Guez, Hubert, Baker, Lai, Bolton, et~al.]{silver2017mastering}
David Silver, Julian Schrittwieser, Karen Simonyan, Ioannis Antonoglou, Aja
  Huang, Arthur Guez, Thomas Hubert, Lucas Baker, Matthew Lai, Adrian Bolton,
  et~al.
\newblock Mastering the game of go without human knowledge.
\newblock \emph{Nature}, 550\penalty0 (7676):\penalty0 354, 2017.

\bibitem[Silver et~al.(2018)Silver, Hubert, Schrittwieser, Antonoglou, Lai,
  Guez, Lanctot, Sifre, Kumaran, Graepel, et~al.]{silver2018general}
David Silver, Thomas Hubert, Julian Schrittwieser, Ioannis Antonoglou, Matthew
  Lai, Arthur Guez, Marc Lanctot, Laurent Sifre, Dharshan Kumaran, Thore
  Graepel, et~al.
\newblock A general reinforcement learning algorithm that masters chess, shogi,
  and go through self-play.
\newblock \emph{Science}, 362\penalty0 (6419):\penalty0 1140--1144, 2018.

\bibitem[Forestier et~al.(2017)Forestier, Mollard, and
  Oudeyer]{forestier2017intrinsically}
S{\'e}bastien Forestier, Yoan Mollard, and Pierre-Yves Oudeyer.
\newblock Intrinsically motivated goal exploration processes with automatic
  curriculum learning.
\newblock \emph{arXiv preprint arXiv:1708.02190}, 2017.

\bibitem[Schmidhuber(2013)]{schmidhuber2013powerplay}
J{\"u}rgen Schmidhuber.
\newblock Powerplay: Training an increasingly general problem solver by
  continually searching for the simplest still unsolvable problem.
\newblock \emph{Frontiers in psychology}, 4:\penalty0 313, 2013.

\bibitem[Standish(2003)]{standish:ijcia03}
Russell~K Standish.
\newblock Open-ended artificial evolution.
\newblock \emph{International Journal of Computational Intelligence and
  Applications}, 3\penalty0 (02):\penalty0 167--175, 2003.

\bibitem[Langdon(2005)]{langdon:metas05}
W.~B. Langdon.
\newblock Pfeiffer -- {A} distributed open-ended evolutionary system.
\newblock In Bruce Edmonds, Nigel Gilbert, Steven Gustafson, David Hales, and
  Natalio Krasnogor, editors, \emph{AISB'05: Proceedings of the Joint Symposium
  on Socially Inspired Computing (METAS 2005)}, pages 7--13, 2005.
\newblock URL
  \url{http://www.cs.ucl.ac.uk/staff/W.Langdon/ftp/papers/wbl_metas2005.pdf}.

\bibitem[Bedau(2008)]{bedau:alife08}
Mark Bedau.
\newblock The arrow of complexity hypothesis (abstract).
\newblock In Seth Bullock, Jason Noble, Richard Watson, and Mark Bedau,
  editors, \emph{Proceedings of the Eleventh International Conference on
  Artificial Life (Alife XI)}, page 750, Cambridge, MA, 2008. MIT Press.
\newblock URL \url{http://www.alifexi.org/papers/ALIFExi-abstracts-0010.pdf}.

\bibitem[Stanley et~al.(2017)Stanley, Lehman, and Soros]{ken:oriley17}
Kenneth~O. Stanley, Joel Lehman, and Lisa Soros.
\newblock Open-endedness: The last grand challenge you’ve never heard of.
\newblock \emph{O'Reilly Online}, 2017.
\newblock URL
  \url{https://www.oreilly.com/ideas/open-endedness-the-last-grand-challenge-youve-never-heard-of}.

\bibitem[Taylor et~al.(2016)Taylor, Bedau, Channon, and et~al.]{taylor:alife16}
Tim Taylor, Mark Bedau, Alastair Channon, and David~Ackley et~al.
\newblock Open-ended evolution: Perspectives from the oee workshop in york.
\newblock \emph{Artificial life}, 22\penalty0 (3):\penalty0 408–423, 2016.

\bibitem[Lehman and Stanley(2008)]{lehman:alife08}
Joel Lehman and Kenneth~O. Stanley.
\newblock Exploiting open-endedness to solve problems through the search for
  novelty.
\newblock In Seth Bullock, Jason Noble, Richard Watson, and Mark Bedau,
  editors, \emph{Proceedings of the Eleventh International Conference on
  Artificial Life (Alife XI)}, Cambridge, MA, 2008. MIT Press.
\newblock URL \url{http://eplex.cs.ucf.edu/papers/lehman_alife08.pdf}.

\bibitem[Graening et~al.(2010)Graening, Aulig, and Olhofer]{graening:ppsn10}
Lars Graening, Nikola Aulig, and Markus Olhofer.
\newblock Towards directed open-ended search by a novelty guided evolution
  strategy.
\newblock In Robert Schaefer, Carlos Cotta, Joanna Ko{\l}odziej, and G\"{u}nter
  Rudolph, editors, \emph{Parallel Problem Solving from Nature -- PPSN XI},
  volume 6239 of \emph{Lecture Notes in Computer Science}, pages 71--80.
  Springer, 2010.
\newblock ISBN 978-3-642-15870-4.

\bibitem[Soros and Stanley(2014)]{soros:alife14}
L.B. Soros and Kenneth~O Stanley.
\newblock Identifying necessary conditions for open-ended evolution through the
  artificial life world of chromaria.
\newblock In \emph{ALIFE 14: The Fourteenth International Conference on the
  Synthesis and Simulation of Living Systems}, pages 793--800, 2014.

\bibitem[Soros et~al.(2016)Soros, Cheney, and Stanley]{soros2:alife14}
L.B. Soros, Nick Cheney, and Kenneth~O Stanley.
\newblock How the strictness of the minimal criterion impacts open-ended
  evolution.
\newblock In \emph{ALIFE 15: The Fifteenth International Conference on the
  Synthesis and Simulation of Living Systems}, pages 208--215, 2016.

\bibitem[Brant and Stanley(2017)]{brant:mcc17}
Jonathan~C. Brant and Kenneth~O. Stanley.
\newblock Minimal criterion coevolution: A new approach to open-ended search.
\newblock In \emph{Proceedings of the 2017 on Genetic and Evolutionary
  Computation Conference (GECCO)}, pages 67--74, 2017.

\bibitem[Ray(1991)]{ray:alife91}
Thomas~S Ray.
\newblock An approach to the synthesis of life.
\newblock In \emph{Artificial Life {II}}, pages 371--408. Addison-Wesley, 1991.

\bibitem[Ficici and Pollack(1998)]{ficici:alife98}
S.G. Ficici and J.B. Pollack.
\newblock {Challenges in coevolutionary learning: Arms-race dynamics,
  open-endedness, and mediocre stable states}.
\newblock \emph{Artificial life VI}, page 238, 1998.

\bibitem[Wiegand et~al.(2001)Wiegand, Liles, and Jong]{wiegand:gecco01}
R.~Paul Wiegand, William~C. Liles, and Kenneth A.~De Jong.
\newblock An empirical analysis of collaboration methods in cooperative
  coevolutionary algorithms.
\newblock In \emph{Proceedings of the 3rd Annual Conference on Genetic and
  Evolutionary Computation}, GECCO'01, pages 1235--1242, San Francisco, CA,
  USA, 2001. Morgan Kaufmann Publishers Inc.
\newblock ISBN 1-55860-774-9.
\newblock URL \url{http://dl.acm.org/citation.cfm?id=2955239.2955458}.

\bibitem[Popovici et~al.(2012)Popovici, Bucci, Wiegand, and
  De~Jong]{popovici:hnc12}
Elena Popovici, Anthony Bucci, R.~Paul Wiegand, and Edwin~D. De~Jong.
\newblock \emph{Coevolutionary Principles}, pages 987--1033.
\newblock Springer Berlin Heidelberg, Berlin, Heidelberg, 2012.
\newblock ISBN 978-3-540-92910-9.
\newblock \doi{10.1007/978-3-540-92910-9_31}.
\newblock URL \url{http://dx.doi.org/10.1007/978-3-540-92910-9_31}.

\bibitem[Bansal et~al.(2018)Bansal, Pachocki, Sidor, Sutskever, and
  Mordatch]{selfplay:arxiv18}
Trapit Bansal, Jakub Pachocki, Szymon Sidor, Ilya Sutskever, and Igor Mordatch.
\newblock Emergent complexity via multi-agent competition.
\newblock \emph{arXiv preprint arXiv:1710.03748}, 2018.

\bibitem[Goodfellow et~al.(2014)Goodfellow, Pouget-Abadie, Mirza, Xu,
  Warde-Farley, Ozair, Courville, and Bengio]{goodfellow_generative_2014}
Ian Goodfellow, Jean Pouget-Abadie, Mehdi Mirza, Bing Xu, David Warde-Farley,
  Sherjil Ozair, Aaron Courville, and Yoshua Bengio.
\newblock Generative adversarial nets.
\newblock In \emph{Advances in neural information processing systems}, pages
  2672--2680, 2014.

\bibitem[Moran and Pollack(2018)]{moran:arxiv18}
Nick Moran and Jordan~B. Pollack.
\newblock Coevolutionary neural population models.
\newblock \emph{CoRR}, abs/1804.04187, 2018.
\newblock URL \url{http://arxiv.org/abs/1804.04187}.

\bibitem[Florensa et~al.(2018)Florensa, Held, Geng, and
  Abbeel]{florensa2018automatic}
Carlos Florensa, David Held, Xinyang Geng, and Pieter Abbeel.
\newblock Automatic goal generation for reinforcement learning agents.
\newblock In \emph{International Conference on Machine Learning}, pages
  1514--1523, 2018.

\bibitem[Florensa et~al.(2017)Florensa, Held, Wulfmeier, Zhang, and
  Abbeel]{pmlr-v78-florensa17a}
Carlos Florensa, David Held, Markus Wulfmeier, Michael Zhang, and Pieter
  Abbeel.
\newblock Reverse curriculum generation for reinforcement learning.
\newblock In \emph{Proceedings of the 1st Annual Conference on Robot Learning},
  pages 482--495, 2017.

\bibitem[Matiisen et~al.(2017)Matiisen, Oliver, Cohen, and
  Schulman]{matiisen2017teacher}
Tambet Matiisen, Avital Oliver, Taco Cohen, and John Schulman.
\newblock Teacher-student curriculum learning.
\newblock \emph{arXiv preprint arXiv:1707.00183}, 2017.

\bibitem[Togelius et~al.(2011)Togelius, Yannakakis, Stanley, and
  Browne]{togelius:ieeetciaig11}
Julian Togelius, Georgios~N Yannakakis, Kenneth~O Stanley, and Cameron Browne.
\newblock Search-based procedural content generation: A taxonomy and survey.
\newblock \emph{{IEEE} Transactions on Computational Intelligence and AI in
  Games}, 3\penalty0 (3):\penalty0 172--186, 2011.

\bibitem[Shaker et~al.(2016)Shaker, Togelius, and Nelson]{shaker2016procedural}
Noor Shaker, Julian Togelius, and Mark~J Nelson.
\newblock \emph{Procedural content generation in games}.
\newblock Springer, 2016.

\bibitem[Justesen et~al.(2018)Justesen, Torrado, Bontrager, Khalifa, Togelius,
  and Risi]{justesen2017procedural}
Niels Justesen, Ruben~Rodriguez Torrado, Philip Bontrager, Ahmed Khalifa,
  Julian Togelius, and Sebastian Risi.
\newblock Illuminating generalization in deep reinforcement learning through
  procedural level generation.
\newblock \emph{arXiv preprint arXiv:1806.10729}, 2018.

\bibitem[Lehman and Stanley(2011{\natexlab{a}})]{lehman:gecco11}
Joel Lehman and Kenneth~O. Stanley.
\newblock Evolving a diversity of virtual creatures through novelty search and
  local competition.
\newblock In \emph{GECCO '11: Proceedings of the 13th annual conference on
  Genetic and evolutionary computation}, pages 211--218, 2011{\natexlab{a}}.

\bibitem[Mouret and Clune(2015)]{mouret2015illuminating}
Jean-Baptiste Mouret and Jeff Clune.
\newblock Illuminating search spaces by mapping elites.
\newblock \emph{arXiv preprint arXiv:1504.04909}, 2015.

\bibitem[Cully et~al.(2015)Cully, Clune, Tarapore, and Mouret]{cully:nature15}
A.~Cully, J.~Clune, D.~Tarapore, and J.-B. Mouret.
\newblock Robots that can adapt like animals.
\newblock \emph{Nature}, 521:\penalty0 503--507, 2015.
\newblock \doi{10.1038/nature14422}.

\bibitem[Nguyen et~al.(2016)Nguyen, Yosinski, and Clune]{nguyen2016innovative}
Anh Nguyen, Jason Yosinski, and Jeff Clune.
\newblock Understanding innovation engines: Automated creativity and improved
  stochastic optimization via deep learning.
\newblock \emph{Evolutionary Computation}, 24\penalty0 (3):\penalty0 545--572,
  2016.

\bibitem[Huizinga et~al.(2016)Huizinga, Mouret, and Clune]{huizinga2016does}
Joost Huizinga, Jean-Baptiste Mouret, and Jeff Clune.
\newblock Does aligning phenotypic and genotypic modularity improve the
  evolution of neural networks?
\newblock In \emph{Proceedings of the 2016 on Genetic and Evolutionary
  Computation Conference (GECCO)}, pages 125--132, 2016.

\bibitem[Huizinga and Clune(2018)]{Joost:arxiv18}
Joost Huizinga and Jeff Clune.
\newblock Evolving multimodal robot behavior via many stepping stones with the
  combinatorial multi-objective evolutionary algorithm.
\newblock \emph{arXiv preprint arXiv:1807.03392}, 2018.

\bibitem[Salimans et~al.(2017)Salimans, Ho, Chen, and Sutskever]{es}
Tim Salimans, Jonathan Ho, Xi~Chen, and Ilya Sutskever.
\newblock Evolution strategies as a scalable alternative to reinforcement
  learning.
\newblock \emph{arXiv preprint arXiv:1703.03864}, 2017.

\bibitem[Lehman and Stanley(2011{\natexlab{b}})]{lehman}
Joel Lehman and Kenneth~O. Stanley.
\newblock Novelty search and the problem with objectives.
\newblock In \emph{Genetic Programming Theory and Practice IX (GPTP 2011)},
  2011{\natexlab{b}}.

\bibitem[Pugh et~al.(2016)Pugh, Soros, and Stanley]{pugh:frontiers16}
Justin~K Pugh, Lisa~B. Soros, and Kenneth~O. Stanley.
\newblock Quality diversity: A new frontier for evolutionary computation.
\newblock 3\penalty0 (40), 2016.
\newblock ISSN 2296-9144.

\bibitem[Krizhevsky et~al.(2012)Krizhevsky, Sutskever, and Hinton]{alexnet2012}
Alex Krizhevsky, Ilya Sutskever, and Geoffrey~E Hinton.
\newblock Imagenet classification with deep convolutional neural networks.
\newblock In \emph{Advances in neural information processing systems}, pages
  1097--1105, 2012.

\bibitem[Eysenbach et~al.(2018)Eysenbach, Gupta, Ibarz, and
  Levine]{eysenbach:arxiv18}
Benjamin Eysenbach, Abhishek Gupta, Julian Ibarz, and Sergey Levine.
\newblock Diversity is all you need: Learning skills without a reward function.
\newblock \emph{arXiv preprint arXiv:1802.06070}, 2018.

\bibitem[Savinov et~al.(2018)Savinov, Raichuk, Marinier, Vincent, Pollefeys,
  Lillicrap, and Gelly]{savinov:arxiv18}
Nikolay Savinov, Anton Raichuk, Raphael Marinier, Damien Vincent, Marc
  Pollefeys, Timothy Lillicrap, and Sylvain Gelly.
\newblock Episodic curiosity through reachability.
\newblock \emph{arXiv preprint arXiv:1810.0227}, 2018.

\bibitem[Lehman and Stanley(2010)]{lehman:gecco10a}
Joel Lehman and Kenneth~O. Stanley.
\newblock Revising the evolutionary computation abstraction: minimal criteria
  novelty search.
\newblock In \emph{Proceedings of the 12th annual conference on Genetic and
  evolutionary computation}, GECCO '10, pages 103--110, New York, NY, USA,
  2010. ACM.
\newblock ISBN 978-1-4503-0072-8.
\newblock \doi{http://doi.acm.org/10.1145/1830483.1830503}.
\newblock URL \url{http://doi.acm.org/10.1145/1830483.1830503}.

\bibitem[Rechenberg(1978)]{rechenberg1978evolutionsstrategien}
Ingo Rechenberg.
\newblock Evolutionsstrategien.
\newblock In \emph{Simulationsmethoden in der Medizin und Biologie}, pages
  83--114. 1978.

\bibitem[Williams(1992)]{williams1992simple}
Ronald~J Williams.
\newblock Simple statistical gradient-following algorithms for connectionist
  reinforcement learning.
\newblock \emph{Machine learning}, 8\penalty0 (3-4):\penalty0 229--256, 1992.

\bibitem[Wierstra et~al.(2008)Wierstra, Schaul, Peters, and
  Schmidhuber]{wierstra2008natural}
Daan Wierstra, Tom Schaul, Jan Peters, and Juergen Schmidhuber.
\newblock Natural evolution strategies.
\newblock In \emph{Evolutionary Computation, 2008.}, pages 3381--3387, 2008.

\bibitem[Sehnke et~al.(2010)Sehnke, Osendorfer, R{\"u}ckstie{\ss}, Graves,
  Peters, and Schmidhuber]{sehnke2010parameter}
Frank Sehnke, Christian Osendorfer, Thomas R{\"u}ckstie{\ss}, Alex Graves, Jan
  Peters, and J{\"u}rgen Schmidhuber.
\newblock Parameter-exploring policy gradients.
\newblock \emph{Neural Networks}, 23\penalty0 (4):\penalty0 551--559, 2010.

\bibitem[Lehman et~al.(2018)Lehman, Chen, Clune, and
  Stanley]{lehman:gecco18esfd}
Joel Lehman, Jay Chen, Jeff Clune, and Kenneth~O. Stanley.
\newblock Es is more than just a traditional finite-difference approximator.
\newblock In \emph{Proceedings of the Genetic and Evolutionary Computation
  Conference}, GECCO '18, pages 450--457, New York, NY, USA, 2018. ACM.
\newblock ISBN 978-1-4503-5618-3.
\newblock \doi{10.1145/3205455.3205474}.
\newblock URL \url{http://doi.acm.org/10.1145/3205455.3205474}.

\bibitem[Zhang et~al.(2017)Zhang, Clune, and Stanley]{zhang:arxiv17}
Xingwen Zhang, Jeff Clune, and Kenneth~O. Stanley.
\newblock On the relationship between the openai evolution strategy and
  stochastic gradient descent.
\newblock \emph{arXiv preprint arXiv:1712.06564}, 2017.

\bibitem[Mnih et~al.(2016)Mnih, Badia, Mirza, Graves, Lillicrap, Harley,
  Silver, and Kavukcuoglu]{a3c}
Volodymyr Mnih, Adria~Puigdomenech Badia, Mehdi Mirza, Alex Graves, Timothy
  Lillicrap, Tim Harley, David Silver, and Koray Kavukcuoglu.
\newblock Asynchronous methods for deep reinforcement learning.
\newblock In \emph{ICML}, pages 1928--1937, 2016.

\bibitem[Conti et~al.(2018)Conti, Madhavan, Petroski~Such, Lehman, Stanley, and
  Clune]{conti:nses_nips2018}
Edoardo Conti, Vashisht Madhavan, Felipe Petroski~Such, Joel Lehman, Kenneth
  Stanley, and Jeff Clune.
\newblock Improving exploration in evolution strategies for deep reinforcement
  learning via a population of novelty-seeking agents.
\newblock In \emph{Advances in Neural Information Processing Systems (NeurIPS)
  31}, pages 5032--5043. 2018.

\bibitem[Stanley and Lehman(2015)]{stanley:book15}
Kenneth~O Stanley and Joel Lehman.
\newblock Why greatness cannot be planned.
\newblock 2015.

\bibitem[Brockman et~al.(2016)Brockman, Cheung, Pettersson, Schneider,
  Schulman, Tang, and Zaremba]{brockman}
Greg Brockman, Vicki Cheung, Ludwig Pettersson, Jonas Schneider, John Schulman,
  Jie Tang, and Wojciech Zaremba.
\newblock Openai gym, 2016.

\bibitem[Ha(2018)]{davidha:arxiv18}
David Ha.
\newblock Reinforcement learning for improving agent design.
\newblock \emph{arXiv preprint arXiv:1810.03779}, 2018.

\bibitem[Kingma and Ba(2014)]{kingma2014adam}
Diederik Kingma and Jimmy Ba.
\newblock Adam: A method for stochastic optimization.
\newblock \emph{arXiv preprint arXiv:1412.6980}, 2014.

\bibitem[Ha(2017)]{dha_esblog}
David Ha.
\newblock Evolving stable strategies.
\newblock \url{http://blog.otoro.net/}, 2017.

\bibitem[Gomez and Miikkulainen(1997)]{gomez:ab97}
Faustino Gomez and Risto Miikkulainen.
\newblock Incremental evolution of complex general behavior.
\newblock \emph{Adaptive Behavior}, 5:\penalty0 317--342, 1997.
\newblock URL \url{gomez:ab97}.

\bibitem[Bengio et~al.(2009)Bengio, Louradour, Collobert, and
  Weston]{bengio2009curriculum}
Yoshua Bengio, J{\'e}r{\^o}me Louradour, Ronan Collobert, and Jason Weston.
\newblock Curriculum learning.
\newblock In \emph{Proceedings of the 26th annual international conference on
  machine learning}, pages 41--48. ACM, 2009.

\bibitem[Karpathy and Van De~Panne(2012)]{karpathy2012curriculum}
Andrej Karpathy and Michiel Van De~Panne.
\newblock Curriculum learning for motor skills.
\newblock In \emph{Canadian Conference on Artificial Intelligence}, pages
  325--330. Springer, 2012.

\bibitem[Heess et~al.(2017{\natexlab{a}})Heess, Sriram, Lemmon, Merel, Wayne,
  Tassa, Erez, Wang, Eslami, Riedmiller, et~al.]{heess2017emergence}
Nicolas Heess, Srinivasan Sriram, Jay Lemmon, Josh Merel, Greg Wayne, Yuval
  Tassa, Tom Erez, Ziyu Wang, Ali Eslami, Martin Riedmiller, et~al.
\newblock Emergence of locomotion behaviours in rich environments.
\newblock \emph{arXiv preprint arXiv:1707.02286}, 2017{\natexlab{a}}.

\bibitem[Klimov(2016)]{OpenAI_bipedal}
O.~Klimov.
\newblock Bipedalwalkerhardcore-v2.
\newblock \url{https://gym.openai.com}, 2016.

\bibitem[Stanley(2007)]{stanley:gpem07}
Kenneth~O. Stanley.
\newblock Compositional pattern producing networks: A novel abstraction of
  development.
\newblock \emph{Genetic Programming and Evolvable Machines Special Issue on
  Developmental Systems}, 8\penalty0 (2):\penalty0 131--162, 2007.

\bibitem[Cheney et~al.(2013)Cheney, MacCurdy, Clune, and
  Lipson]{cheney:gecco13}
N.~Cheney, R.~MacCurdy, J.~Clune, and H.~Lipson.
\newblock Unshackling evolution: evolving soft robots with multiple materials
  and a powerful generative encoding.
\newblock In \emph{Proceedings of the Genetic and Evolutionary Computation
  Conference (GECCO-2013)}, New York, NY, 2013. ACM Press.

\bibitem[Stanley et~al.(2009)Stanley, D'Ambrosio, and Gauci]{stanley:alife09}
Kenneth~O. Stanley, David~B. D'Ambrosio, and Jason Gauci.
\newblock A hypercube-based indirect encoding for evolving large-scale neural
  networks.
\newblock \emph{Artificial Life}, 15\penalty0 (2):\penalty0 185--212, 2009.
\newblock URL
  \url{http://eplex.cs.ucf.edu/publications/2009/stanley.alife09.html}.

\bibitem[Clune et~al.(2011)Clune, Stanley, Pennock, and Ofria]{clune:ieeetec11}
Jeff Clune, Kenneth~O. Stanley, Robert~T. Pennock, and Charles Ofria.
\newblock On the performance of indirect encoding across the continuum of
  regularity.
\newblock \emph{IEEE Transactions on Evolutionary Computation}, 2011.

\bibitem[Clune et~al.(2009)Clune, Beckmann, Ofria, and Pennock]{clune:cec09}
Jeff Clune, Benjamin~E. Beckmann, Charles Ofria, and Robert~T. Pennock.
\newblock Evolving coordinated quadruped gaits with the {HyperNEAT} generative
  encoding.
\newblock In \emph{Proceedings of the IEEE Congress on Evolutionary Computation
  (CEC-2009) Special Session on Evolutionary Robotics}, Piscataway, NJ, USA,
  2009. IEEE Press.

\bibitem[Yosinski et~al.(2011)Yosinski, Clune, Hidalgo, Nguyen, Zagal, and
  Lipson]{yosinski:ecal11}
Jason Yosinski, Jeff Clune, Diana Hidalgo, Sarah Nguyen, Juan~Cristobal Zagal,
  and Hod Lipson.
\newblock Evolving robot gaits in hardware: the hyperneat generative encoding
  vs. parameter optimization.
\newblock In \emph{Proceedings of the 20th European Conference on Artificial
  Life}, August 2011.

\bibitem[Schulman et~al.(2015)Schulman, Levine, Abbeel, Jordan, and
  Moritz]{schulman2015trust}
John Schulman, Sergey Levine, Pieter Abbeel, Michael Jordan, and Philipp
  Moritz.
\newblock Trust region policy optimization.
\newblock In \emph{International Conference on Machine Learning}, pages
  1889--1897, 2015.

\bibitem[Schulman et~al.(2017)Schulman, Wolski, Dhariwal, Radford, and
  Klimov.]{ppo:arxiv17}
J.~Schulman, F.~Wolski, P.~Dhariwal, A.~Radford, and O.~Klimov.
\newblock Proximal policy optimization algorithms.
\newblock \emph{arXiv preprint arXiv:1707.06347}, 2017.

\bibitem[Gupta et~al.(2018)Gupta, Eysenbach, Finn, and
  Levine]{gupta2018unsupervised}
Abhishek Gupta, Benjamin Eysenbach, Chelsea Finn, and Sergey Levine.
\newblock Unsupervised meta-learning for reinforcement learning.
\newblock \emph{arXiv}, 2018.
\newblock URL \url{http://arxiv.org/abs/1806.04640}.

\bibitem[Heess et~al.(2017{\natexlab{b}})Heess, TB, Sriram, Lemmon, Merel,
  Wayne, Tassa, Erez, Wang, Eslami, Riedmiller, and Silver]{heess:arxiv17}
Nicolas Heess, Dhruva TB, Srinivasan Sriram, Jay Lemmon, Josh Merel, Greg
  Wayne, Yuval Tassa, Tom Erez, Ziyu Wang, S.~M.~Ali Eslami, Martin~A.
  Riedmiller, and David Silver.
\newblock Emergence of locomotion behaviours in rich environments.
\newblock \emph{CoRR}, abs/1707.02286, 2017{\natexlab{b}}.
\newblock URL \url{http://arxiv.org/abs/1707.02286}.

\bibitem[Lehman and Stanley(2011{\natexlab{c}})]{lehman:ecj11}
Joel Lehman and Kenneth~O. Stanley.
\newblock Abandoning objectives: Evolution through the search for novelty
  alone.
\newblock \emph{Evolutionary Computation}, 19\penalty0 (2):\penalty0 189--223,
  2011{\natexlab{c}}.
\newblock URL
  \url{http://www.mitpressjournals.org/doi/pdf/10.1162/EVCO_a_00025}.

\end{thebibliography}

\newpage
\section{Supplemental Information}\label{supplemental}

The first section in this supplement details the procedure in POET for generating and choosing new environments, which is followed by an explanation of how POET attempts to transfers agents from one environment to another.

\subsection{Producing New Environments}\label{mutation_section}


New environments are produced by copying the current environments and making random changes to them 
(in the language of evolutionary algorithms, the environments \emph{reproduce} through \emph{mutation}). 
Algorithm \ref{alg:mutate_niches}
formalizes the single mutation step, given a list of environment-agent pairs. The primary logic in this process is: (1) Each environment evaluates its paired agent. When the reward obtained by the agent satisfies a predefined reproduction eligibility condition for the environment (denoted as ELIGIBLE\_TO\_REPRODUCE()), the environment is added to the list of parent environments.
(2) Once the list of eligible parents is complete, mutations of the parent environments return a list of child environments (through the function ENV\_REPRODUCE()). Each child environment's paired agent is initialized as a clone of its parent environment's paired agent.
(3) Each child environment $E^{\textrm{child}}(\cdot)$, together with its paired agent $\theta^{\textrm{child}}$, is checked against a predefined minimal criterion (reminiscent of the minimal criterion in the MCC algorithm \cite{brant:mcc17}) and only those satisfying the minimal criterion are kept (denoted as MC\_SATISFIED()), ensuring that potential new environments are not too trivial and not too hard (which would lead to little progress). 
(4) All remaining child environments are then ranked based on their novelty (detailed shortly), 
which is calculated with respect to the current active population of environments and an \emph{archive} of all previously active environments (i.e.\ those that were allowed to enter the population previously).
(5) A transfer attempt is then performed to see whether an agent from a different environment could lead to the best performance in a new child environment.  The logic of transfer is described in the next section (Section \ref{Transfer Overview}).  Each pair is then checked against the minimal criterion again in case some pairs no longer satisfy the minimal criterion after the transfer (e.g.\ if 
an environment is exposed as too easy). 

To limit the usage of computational resources, (6) POET caps the number of child environments that can be created during one iteration of POET's main loop.
The number of new environments that can be admitted is $\texttt{max\_admitted}$ and the maximum number of active environments at any time is $\texttt{capacity}$. If adding new environments exceeds the system's capacity, the oldest environments and their paired agents in the system are removed to make room (also as in the MCC algorithm).

The \emph{encoding} of an environment in POET is its underlying description in the system, such as a vector of numbers that specify the distribution of obstacles in an obstacle course.
As mentioned in step (4) above, POET calculates the novelty of the child environments.  This novelty calculation is based on the encoding. 
Child environments are ranked based on their novelty.
For readers familiar with the \emph{novelty search} algorithm \cite{lehman:ecj11}, it is important to note here that the encoding of an environment is in effect a genetic representation, which means that in POET in effect novelty is a \emph{genetic} diversity measure.  That way, POET does not in principle require a behavior characterization (BC) \cite{lehman:ecj11} to be devised for environments, though doing so is also not precluded.

Measuring novelty introduces an explicit pressure for diversity among the environments by encouraging newly admitted environments to have notably different encodings than those previously admitted.
Given an environment $E$ with an encoding denoted as $e(E)$, and a list $L$
containing the encodings of all the environments currently in the population plus those previously admitted but no longer active (which are stored in an \emph{archive}), 
the novelty of $E$, denoted as $N(E, L)$, is then computed by selecting the k-nearest neighbors of $e(E)$ from $L$ and computing the average distance between them:
\begin{gather*}
\scalebox{1.}{$N(e(E), L) = \frac{1}{\vert{S}\vert}\sum_{j\in{S}}\vert\vert{e(E) - e(E_{j})}\vert\vert_{2}$}\\
\scalebox{1.}{$S = \textrm{kNN}(e(E), L)$}\\
\scalebox{1.}{$
= \{e(E_{1}), e(E_{2}), ..., e(E_{k})\}$}
\end{gather*}
Note that in this work the distance between genetic encodings of environments is calculated with an $L2$-norm and the number of nearest neighbors $k$ in the calculation of novelty is 5.


\begin{algorithm}[htp]
\caption{MUTATE\_ENVS}
\label{alg:mutate_niches}
\begin{algorithmic}[1]
   \STATE {\bfseries Input:} list of environment(E)-agent(A) pairs $\texttt{EA\_list}$,
   each entry is a pair $(E(\cdot), \theta)$
   \STATE {\bfseries Parameters:} maximum number of children per reproduction \texttt{max\_children}, maximum number of children admitted per reproduction \texttt{max\_admitted}, maximum number of active environments \texttt{capacity}
   \STATE {\bfseries Initialize:} Set \texttt{parent\_list} to empty
   \STATE $M = \textrm{len}(\texttt{EA\_list})$   
   \FOR{$m=1$ {\bfseries to} $M$}
   \STATE $E^{m}(\cdot)$, $\theta^{m}$ = $\texttt{EA\_list}$[m]
   \IF{ELIGIBLE\_TO\_REPRODUCE$(E^{m}(\cdot), \theta^{m})$}
   \STATE add $(E^{m}(\cdot), \theta^{m})$ to \texttt{parent\_list}
   \ENDIF   
   \ENDFOR 
   \STATE $\texttt{child\_list} = $ ENV\_REPRODUCE$(\texttt{parent\_list}, \texttt{max\_children})$
   \STATE $\texttt{child\_list} = $ MC\_SATISFIED$(\texttt{child\_list})$
   \STATE $\texttt{child\_list} = $ RANK\_BY\_NOVELTY$(\texttt{child\_list})$
   \STATE $\texttt{admitted} = 0$
   \FOR{$E^{\textrm{child}}(\cdot), \theta^{\textrm{child}} \in \texttt{child\_list}$}
   \STATE $\theta^{\textrm{child}} = $ EVALUATE\_CANDIDATES$(\theta^{1}$, $\theta^{2}$, \ldots, $\theta^{M}$, $E^{\textrm{child}}(\cdot)$, $\alpha$, $\sigma))$
   \IF{MC\_SATISFIED$([E^{\textrm{child}}(\cdot), \theta^{\textrm{child}}])$} 
   \STATE add $(E^{\textrm{child}}(\cdot), \theta^{\textrm{child}})$ to $\texttt{EA\_list}$
   \STATE $\texttt{admitted} = \texttt{admitted} + 1$
   \STATE {\bfseries if} $\texttt{admitted} >= \texttt{max\_admitted}$ {\bfseries then} break {\bfseries end if}
   \ENDIF
   \ENDFOR
   \STATE $M = \textrm{len}(\texttt{EA\_list})$
   \IF{$M > \texttt{capacity}$}
   \STATE $\texttt{num\_removals} = M - \texttt{capacity}$
   \STATE REMOVE\_OLDEST$(\texttt{EA\_list}, \texttt{num\_removals})$ 
   \ENDIF
   \STATE {\bfseries Return:} $\texttt{EA\_list}$

\end{algorithmic}
\end{algorithm}

\subsection{ES-Based Transfer} \label{Transfer Overview}
POET adopts a transfer mechanism that allows one agent (the \emph{emigrant})
to attempt to transfer its learned capabilities to 
a \emph{target environment}, possibly replacing the current target environment's agent. This transfer mechanism reflects the idea that skills gained when interacting with one environment might serve as a stepping stone to improved 
performance when interacting with another environment.  We do not know at any given time which environment's mastery might boost another's, but we can always perform \emph{experiments} to see whether such a transfer might improve performance in the target environment. As results support, the richness of simultaneous transfer attempts in a divergent coevolving system can lead to a succession of breakthroughs that would not be possible through a simple monotonic succession of improvements on a single environment.




Algorithm \ref{alg:es_transfer} illustrates the function EVALUATE\_CANDIDATES() that facilitates this transfer process. Given a list of $M$ input candidate agents, denoted as $\theta^{1}$, $\theta^{2}$, \ldots, $\theta^{M}$, and a target environment $E(\cdot)$, this function first concatenates the list with \emph{proposals}, which are calculated by optimizing each of $\theta^{1}$, $\theta^{2}$, \ldots, $\theta^{M}$ in $E(\cdot)$ with one ES step. Then each agent in the augmented list (the original $M$ agents and each of their $M$ proposals) is evaluated in $E(\cdot)$ and the one with the highest reward is returned. As illustrated in Algorithm \ref{alg:poet}, the agent returned from this function will replace the agent currently paired with $E(\cdot)$ if the former performs better than the latter in $E(\cdot)$. We refer to the replacement as \emph{transfer}, and there are two types: the situation when the replacing agent is from the input candidate list (i.e. without an optimization step on $E(\cdot)$) is referred to as a \emph{direct transfer}, while if the replacing agent is one of the proposals, it is called a \emph{proposal transfer}. 

\begin{algorithm}[htp]
\caption{EVALUATE\_CANDIDATES}
\label{alg:es_transfer}
\begin{algorithmic}[1]
   \STATE {\bfseries Input:} candidate agents denoted by their policy parameter vectors $\theta^{1}$, $\theta^{2}$, \ldots, $\theta^{M}$, target environment $E(\cdot)$,  learning rate $\alpha$, noise standard deviation $\sigma$
   \STATE {\bfseries Initialize:} Set list $C$ empty 
   \FOR{$m=1$ {\bfseries to} $M$}
   \STATE Add $\theta^{m}$ to $C$
   \STATE Add $(\theta^{m} +$ ES\_STEP$(\theta^{m}, E(\cdot), \alpha, \sigma)) $ to $C$
   \ENDFOR
   \STATE {\bfseries Return:} $\argmax_{\theta \in \mathcal{C}} E(\theta) $

\end{algorithmic}
\end{algorithm}

\end{document}